\documentclass[sigconf]{acmart}

\usepackage{caption}
\usepackage{subcaption}
\usepackage{multirow}
\usepackage{bm}
\usepackage{comment}
\usepackage{natbib}

\usepackage[bb=dsserif]{mathalpha}

\usepackage{amsmath}
\usepackage{mathtools}
\usepackage{amsthm}

\newcommand{\ie}{\textit{i}.\textit{e}.,}
\newcommand{\eg}{\textit{e}.\textit{g}.,}
\newcommand{\etc}{\textit{etc}.}
\DeclarePairedDelimiter{\norm}{\lVert}{\rVert}
\newcommand{\romsm}[1]{\lowercase\expandafter{\romannumeral #1\relax}}

\DeclareMathOperator*{\argmin}{arg\,min}
\newcommand{\sota}{state-of-the-art}
\renewcommand{\eqref}[1]{Eq.\,(\ref{#1})}

\AtBeginDocument{%
  }

\copyrightyear{2024}
\acmYear{2024}
\setcopyright{acmlicensed}\acmConference[CIKM '24]{Proceedings of the 33rd
ACM International Conference on Information and Knowledge
Management}{October 21--25, 2024}{Boise, ID, USA}
\acmBooktitle{Proceedings of the 33rd ACM International Conference on
Information and Knowledge Management (CIKM '24), October 21--25, 2024,
Boise, ID, USA}
\acmDOI{10.1145/3627673.3679750}
\acmISBN{979-8-4007-0436-9/24/10}

\begin{document}

\title{Towards Robust Vision Transformer via Masked Adaptive Ensemble}


\author{Fudong Lin}
\affiliation{
  \institution{University of Delaware}
  \city{Newark}
  \state{DE}
  \country{USA}
}

\author{Jiadong Lou}
\affiliation{
  \institution{University of Delaware}
  \city{Newark}
  \state{DE}
  \country{USA}
}

\author{Xu Yuan}
  \authornote{Corresponding author: Dr. Xu Yuan (xyuan@udel.edu)}
  \affiliation{%
    \institution{University of Delaware}
    \city{Newark}
    \state{DE}
    \country{USA}
}

\author{Nian-Feng Tzeng} 
\affiliation{
  \institution{University of Louisiana at Lafeyette}
  \city{Lafeyette}
  \state{LA}
  \country{USA}
}

\renewcommand{\shortauthors}{Fudong Lin, Jiadong Lou, Xu Yuan, and Nian-Feng Tzeng}

\begin{abstract}
    Adversarial training (AT) can help improve the robustness of Vision Transformers (ViT) against adversarial attacks by intentionally injecting adversarial examples into the training data.
    However, this way of adversarial injection inevitably incurs standard accuracy degradation to some extent, thereby calling for a trade-off between standard accuracy and adversarial robustness.
    Besides, the prominent AT solutions are still vulnerable to adaptive attacks.
    %
    To tackle such shortcomings,  this paper proposes a novel ViT architecture, including a detector and a classifier bridged by our newly developed adaptive ensemble.
    Specifically, we empirically discover that detecting adversarial examples can benefit from the Guided Backpropagation technique.
    Driven by this discovery, a novel Multi-head Self-Attention (MSA) mechanism is introduced for enhancing our detector to sniff adversarial examples.
    Then, a classifier with two encoders is employed for extracting visual representations respectively from clean images and adversarial examples, 
    with our adaptive ensemble to adaptively adjust the proportion of visual representations from the two encoders for accurate classification.
    This design enables our ViT architecture to achieve a better trade-off between standard accuracy and adversarial robustness.
    Besides, the adaptive ensemble technique allows us to mask off a random subset of image patches within input data,
    boosting our ViT's robustness against adaptive attacks, while maintaining high standard accuracy.
    Experimental results exhibit that 
    our ViT architecture, on CIFAR-10,
    achieves the best standard accuracy and adversarial robustness of $90.3 \%$ and $49.8 \%$, respectively.
\end{abstract}


\maketitle

\section{Introduction}\label{sec:intro}

The Vision Transformers (ViT) architecture has demonstrated impressive capabilities in a wide range of vision tasks, 
including image and video classification~\cite{dosovitskiy:iclr21:vit,touvron:icml21:deit,arnab:iccv21:vivit}, 
dense prediction tasks~\cite{liu:iccv21:swin,wang:iccv21:pvt,wang:cvm22:pvtv2}, 
self-supervised learning~\cite{bao:iclr22:beit,he:cvpr22:mae,tong:nips22:video_mae},
among others~\cite{heo:iccv21:pit,fan:iccv21:mvit,li:cvpr22:mvit2,caron:iccv21:dino,yuan:iccv21:t2tvit,lyu:amia22:attn,fudong:ecml23:storm,lyu:naacl22:attn,fudong:iccv23:mmst_vit,ruan2024infostgcan,he2023robust,ruan2024infostgcan,fudong:kdd24:cropnet}.
However, similar to Convolutional Neural Networks (CNNs)~\cite{alex:nips12:alex_net,simonyan:iclr15:vgg,he2016resnet,zhu2021twitter,fudong:ijcai21:oversampling,ruan2022learning,song2024eta,fudong:cikm22:cascade_vae,hu2023artificial,han2024bundledslam,ruan2023causal},
the ViT architecture 
is vulnerable to adversarial attacks
\cite{goodfellow:iclr15:fgsm,papernot2016JSMA,moosavi:cvpr16:deepfool,carlini:sp17:cw,dong2020greedyfool,pintor:nips21:fmn,zhu2021homotopy}
achieved by maliciously altering clean images within a small distance, 
leading to incorrect predictions with high confidence.
This vulnerability hinders the adoption of ViT in critical domains
such as healthcare, 
finances, \etc\

So far, adversarial training (AT) methods~\cite{madry:iclr18:pgd,zhang:icml19:trades,jia:cvpr22:las,wong:iclr20:fast,andriushchenko:nips20:fgsm-rs,kim:aaai21:single-step,li:cvpr22:sub-at,wang:icml22:remove-batch_norm,ruan2023causal} 
are widely accepted as the most effective mechanisms for improving ViT's robustness against adversarial attacks,
by intentionally injecting adversarial examples into the training data.
Unfortunately, existing AT solutions struggle with two limitations.
First, they suffer from a trade-off between standard accuracy (\ie\ the accuracy on clean images) 
and adversarial robustness (\ie\ the accuracy on adversarial examples),
with improved robustness while yielding non-negligible standard accuracy degradation.
Second, these solutions are not effective against 
adaptive attacks~\cite{croce:icml20:autoattack,tramer:nips20:a2,yao:nips21:auto-adaptive-attack,liu:cvpr22:pf_a3},
\ie\ a category of adversarial attacks capable of exploiting the weak points of defense methods to adaptively adjust their attack strategies.
Hence, it calls for the exploration of enhancing ViT's robustness against adaptive attacks.

One potential direction to tackle the trade-off between standard accuracy and 
adversarial robustness is the \textit{detection/rejection} mechanism. 
This involves training an additional detector to identify and reject malicious input data, 
with several solutions 
proposed in the literature~\cite{roth:icml19:odds,ma:ndss19:nic,yin:iclr20:gat,raghuram:icml21:jtla,tramer:icml22:hard-detect}. 
However, these detection techniques have limited effectiveness against adaptive attacks and cannot be applied to scenarios involving natural adversarial examples, 
as reported in a prior study~\cite{hendrycks:cvpr21:natural-ae}. 
Hence, it is crucial to develop novel solutions that can address limitations associated with the aforementioned direction 
and are suitable for a wide range of scenarios.

In this work, we aim to boost the robustness of ViT against adaptive attacks 
in a more general and challenging scenario where malicious inputs cannot be rejected.
Such a scenario is common to several critical application domains, such as autonomous driving, 
where the system must correctly recognize a road sign even if it has been maliciously crafted. 
To this end, we propose a novel ViT architecture consisting of a detector and a classifier,
connected by a newly developed \textit{adaptive ensemble}.
After adversarially trained by One-step Least-Likely Adversarial Training, 
our proposed ViT architecture can withstand adaptive attacks 
while incurring only a negligible standard accuracy degradation. 

In essence, our detector 
incorporates two innovative designs to make adversarial examples more noticeable.
First, based on our empirical observations,
we introduce a novel Multi-head Self-Attention (MSA) mechanism~\cite{vaswani2017attention} to expose adversarial perturbation
by Guided Backpropagation~\cite{springenberg:iclrw15:guided}.
Second, the Soft-Nearest Neighbors Loss (SNN Loss)~\cite{salakhutdinov:aistats17:snn,frosst:icml19:snn}
is tailored to push adversarial examples away from their corresponding clean images.
Our detector thus can effectively sniff adaptive attack-generated adversarial examples.
On the other hand, our classifier's adversarial training involves two stages: pre-training and fine-tuning. 
During the pre-training stage, our classifier 
utilizes one clean encoder, one adversarial encoder, and one decoder to jointly learn high-quality visual representations 
and encourage pairwise similarity between a clean image and its adversarial example. 
Here, we extend Masked Autoencoders~(MAE)~\cite{he:cvpr22:mae} to facilitate adversarial training through a new design. 
Specifically, we reconstruct images from one pair of a masked clean image and its masked adversarial example, for representation learning,
with a contrastive loss on a pair of visual representations to encourage similarity. 
%
%
In the fine-tuning stage, we discard the decoder and freeze the weights in the well-trained detector and two encoders,
with a newly developed \textit{adaptive ensemble} to bridge the detector and the two encoders, 
for fine-tuning an MLP (Multi-layer Perceptron) for accurate classification. 
%
Our adaptive ensemble also masks off a random subset of image patches within the input, 
enabling our approach to mitigate adversarial effects when encountering malicious inputs.
Extensive experimental results on three popular benchmarks
demonstrate that our approach outperforms state-of-the-art
adversarial training techniques in terms of both standard accuracy and adversarial robustness.

\section{Related Work}\label{sec:rw}

\par\smallskip\noindent
{\bf Detection Mechanisms.} 
Detecting adversarial examples~(AEs) and then rejecting them (\ie\ detection/rejection mechanism) 
can improve the model's robustness against adversarial attacks.
That is, the input will be rejected if the detector classifies it as an adversarial example.
Popular detection techniques include
Odds~\cite{roth:icml19:odds}, which considers the difference between clean images and AEs in terms of log-odds; 
NIC~\cite{ma:ndss19:nic}, which checks channel invariants within deep neural networks (DNNs); 
GAT~\cite{yin:iclr20:gat}, which resorts to multiple binary classifiers; 
JTLA~\cite{raghuram:icml21:jtla}, which proposes a detection framework 
by employing internal layer representations,
among others~\cite{lee:nips18:simple-det,fu2022feature,yang:aaai20:ML-LOO,sheikholeslami:iclr21:provably-det,fu2023differential}.
Unfortunately, existing detection methods are typically ineffective in 
defending against adaptive attacks. 
Besides, the detection/rejection mechanism 
cannot be generalized to domains where natural adversarial examples exist.
Our work differs from previous solutions in two aspects.
First, we introduce a novel Multi-head Self-Attention (MSA) mechanism by using the Guided Backpropagation technique, 
which can largely expose adversarial perturbations.
Second, we incorporate the Soft-Nearest Neighbors (SNN) loss to maximize the differences between clean images and adversarial examples.
These innovative designs enable our detector to effectively defend against adaptive attacks.
Moreover, our newly developed adaptive ensemble further enhances our detector,
empowering it to be applied to scenarios where rejecting input images is not allowed.

\par\smallskip\noindent
{\bf Adversarial Training Approaches.} 
Adversarial training (AT) aims to improve the model's robustness against adversarial attacks
by intentionally injecting adversarial examples into the training data.
For example, PGD-AT~\cite{madry:iclr18:pgd} 
proposes a multi-step attack to find the worst case of training data,
TRADES~\cite{zhang:icml19:trades} addresses the limitation of PGD-AT by utilizing theoretically sound classification-calibrated loss,
EAT~\cite{tramer:iclr18:eat} uses an ensemble of different DNNs to produce the threat model,
FAT~\cite{wong:iclr20:fast} reduces the computational overhead of AT by utilizing FGSM attack with the random initialization,
LAS-AWP~\cite{jia:cvpr22:las} boosts AT with a learnable attack strategy,
Sub-AT~\cite{li:cvpr22:sub-at} constrains AT in a well-designed subspace,
and many others~\cite{shafahi:nips19:Free-AT,andriushchenko:nips20:fgsm-rs,chen2020more,kim:aaai21:single-step,min2021curious,wang:icml22:remove-batch_norm,liu2022evil,he2023robust_agent,liu2023slowlidar,he:iros23:robust_electric,wu2023bottrinet,wu2023botshape,han:cvpr23:riatig,zhuang2022defending,zhuang2022robust,fu2023mitigating}.
However, prior ATs suffer from the dilemma of balancing the trade-off between standard accuracy and adversarial robustness.
Besides, their improved robustness is vulnerable to adaptive attacks.
%
In contrast,
our work introduces a ViT architecture consisting of a detector and a classifier,
connected by a newly developed adaptive ensemble,
able to boost 
AT to defend against adaptive attacks.
Meanwhile, it lowers the standard accuracy degradation
by employing two encoders for extracting visual representations 
respectively from clean images and adversarial examples,
empowering our ViT architecture to enjoy a better trade-off between 
accuracy and robustness.

\section{Preliminary: One-step Least-Likely Adversarial Training} 
\label{sec:preliminary}

%
Adversarial training~(AT) improves the model's robustness against adversarial attacks
by feeding adversarial examples into the training set.
Given a model $f$ with parameters $\bm{\theta}$,
a dataset with $N$ samples, 
\ie\ $\mathbb{X} = \{(\bm{x}_{i}, y_{i}) ~|~ i \in \{1,2, \dots, N\}\}$,
the cross-entropy loss function $\mathcal{L}$,
and a threat model $\bm{\Delta}$,
AT aims to solve
the following inner-maximization problem
and outer-minimization problem, \ie\
\begin{equation} \label{eq:at}
    \min_{\bm{\theta}} \sum_{i}^{N} \max_{\bm{\delta} \in \bm{\Delta}} 
    \mathcal{L} (f_{\bm{\theta}} (\bm{x}_{i} + \bm{\delta}), ~y_{i} ),
\end{equation}
where the inner problem aims to find the worst-case training data for the given model, 
and the outer problem aims to improve the model's performance on such data.
Recently, one-step Fast Adversarial Training (FAT)~\cite{wong:iclr20:fast} is popular 
due to its computational efficiency.
FAT sets the threat model under a small and $l_{\infty}$ constraint $\epsilon$,
\ie\ $\bm{\Delta} = \{ \bm{\delta}:\norm{\bm{\delta}}_{\infty} \leq \epsilon \}$,
by performing Fast Gradient Sign Method~(FGSM)~\cite{goodfellow:iclr15:fgsm}
with the random initialization, \ie\
\begin{equation} \label{eq:fat}
    \begin{gathered} 
    \bm{\delta} = \text{Uniform} (-\epsilon, \epsilon)
    + \epsilon \cdot \textrm{sign} (\nabla_{\bm{x}}~ \mathcal{L} (f_{\bm{\theta}}(\bm{x}_{i}), ~y_{i})), \\
    \bm{\delta} = \max(\min(\bm{\delta}, \epsilon), -\epsilon), 
    \end{gathered}
\end{equation}    
where $\textit{Uniform}$ denotes the uniform distribution 
and $\textit{sign}$ is the sign function. 
Notably, the second row in Eq.~(\ref{eq:fat}) serves to
project the perturbation $\bm{\delta}$ back into the $l_{\infty}$ ball
around the data $\bm{x}_{i}$.

To find the worst-case adversarial examples,
we extend FAT by performing the least-likely targeted attacks,
inspired by prior studies~\cite{kurakin2017ll_attack,tramer:iclr18:eat}.
That is, given an input $\bm{x}_{i}$, we perform targeted FGSM
by setting the targeted label as its least-likely class, 
\ie\ $y_{i}^{ll} = \argmin f_{\bm{\theta}}(\bm{x}_{i})$,
arriving at,
\begin{equation} \label{eq:fat-ll}
    \begin{gathered}
        \bm{\delta} = \text{Uniform} (-\epsilon, \epsilon)
        + \epsilon \cdot \textrm{sign} (\nabla_{\bm{x}}~ \mathcal{L} (f_{\bm{\theta}}(\bm{x}_{i}), ~y_{i}^{ll})), \\
        \bm{\delta} = \max(\min(\bm{\delta}, \epsilon), -\epsilon), 
    \end{gathered}  
\end{equation}
Our one-step least-likely adversarial training is to utilize Eq.(\ref{eq:fat-ll}) to produce the threat model.

\section{Our Approaches} \label{sec:our_approach}

\subsection{Problem Statement} \label{sec:ps}
We consider a set of $N$ samples, \ie\ 
$ \mathbb{X}$ = $\{(\bm{x}_{i}, ~y_{i}) ~|~ i \in \{1,2, \dots, N\} \}$, 
where $\bm{x} \in \mathbb{R}^{H \times W \times C_{H}}$ 
is an input image with the resolution of $(H, W)$
and the channel count of $C_{H}$,
and $y \in [C]$ denotes its label.
For notational convenience, we let $d = H \times W \times C_{H}$. 
A classifier is a function $f_{\bm{\theta}}$:
$\mathbb{R}^{d} \rightarrow [C]$, parameterized by a neural network.
We consider two types of inputs,
\ie\ a clean image $ \bm{x}^{\text{cln}}$ sampled from the standard distribution 
$\mathcal{D}_{\textrm{std}}$
and an adversarial example $ \bm{x}^{\text{adv}}$
sampled from the adversarial distribution $\mathcal{D}_{\textrm{adv}}$.
We assume $\mathcal{D}_{\textrm{std}}$ and $\mathcal{D}_{\textrm{adv}}$ 
follow different distributions.
The clean image $\bm{x}^{\text{cln}}$ itself 
or its augmented variant can be the input,
while the adversarial example $\bm{x}^{\text{adv}}$ is a malicious version of $\bm{x}$ within a small distance. 
That is, for some metric $d$, we have $d(\bm{x}, \bm{x^{\textrm{adv}}}) \leq \epsilon $,
but $\bm{x}^{\textrm{adv}}$ can mislead conventional classifiers.
Parameterized by another neural network,
a detector $g_{\bm{\phi}}$ is to tell whether an input image is a clean image or not,
\ie\ $g_{\bm{\phi}}$ : $\mathbb{R}^{d} \rightarrow \{\pm 1 \}$,
where $+1$ and $-1$ indicate a clean image and an adversarial example, respectively.
The binary indicator function $\mathbb{1}_{\{ \cdot \}}$ is $1$
if both the detector $g_{\bm{\phi}}$ and the classifier $f_{\bm{\theta}}$
make correct predictions.
%
We follow previous studies~\cite{madry:iclr18:pgd,zhang:icml19:trades} 
by referring standard accuracy, 
and adversarial robustness, 
as classification accuracy on clean images and adversarial examples, respectively.

\subsection{Detector} \label{sec:det}
Parameterized by a neural network with parameters $\bm{\phi}$,
the detector  $g_{\bm{\phi}}$ : $\mathbb{R}^{d} \rightarrow \{ \pm 1 \}$
is to determine whether the input is a clean image or not,
where $+1$ and $-1$ respectively represent
a clean image and an adversarial example, \ie\
\begin{equation} \label{eq:det-obj}
  g_{\bm{\phi}} (\bm{x}) = \begin{cases}
    +1,    & \text{if $\bm{x}$ is a clean image} \\
    -1,  & \text{otherwise}.
  \end{cases}
\end{equation}
Aiming to generalize the robust model to critical domains~(\eg\ autonomous driving),
the input will not be rejected in this work.
Instead, we have modified it to output 
an estimated probability of $p \in [0, 1]$ for clean images 
and $1-p$ for adversarial examples.

\begin{figure}[!t]
  \centering
  \captionsetup[subfigure]{justification=centering}
  \begin{subfigure}[t]{0.15\textwidth}
      \centering
       \includegraphics[width=\textwidth]{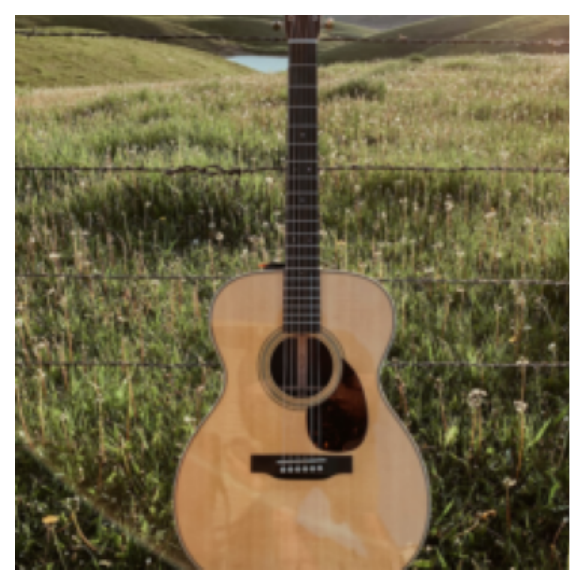}
         \vspace{-1em}
      \caption{Clean image 
      }
      \label{fig:method-det-cln}
  \end{subfigure}
  \centering
   \begin{subfigure}[t]{0.15\textwidth}
      \centering
       \includegraphics[width=\textwidth]{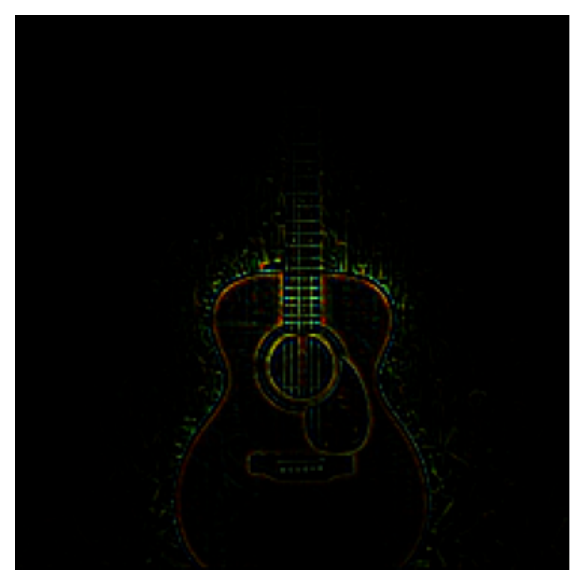}
         \vspace{-1em}
      \caption{Guided Grad-CAM 
      }
      \label{fig:method-det-grad-cam-cln}
  \end{subfigure}
  \centering
  \begin{subfigure}[t]{0.15\textwidth}
     \centering
      \includegraphics[width=\textwidth]{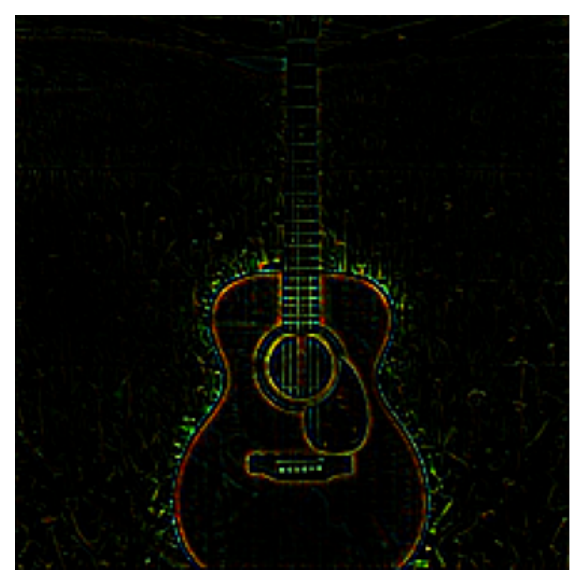}
        \vspace{-1em}
     \caption{ Guided Backprop 
     }
     \label{fig:method-det-guided-cln}
  \end{subfigure}

  \centering
  \captionsetup[subfigure]{justification=centering}
  \begin{subfigure}[t]{0.15\textwidth}
      \centering
       \includegraphics[width=\textwidth]{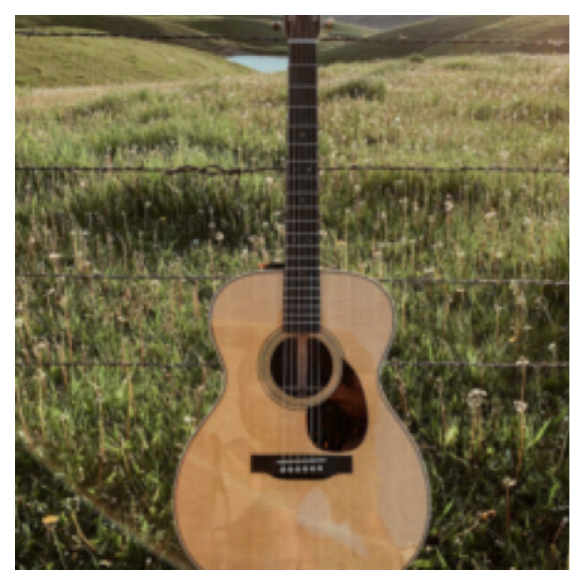}
         \vspace{-1em}
      \caption{Adversarial example
      }
      \label{fig:method-det-adv}
  \end{subfigure}
  \centering
   \begin{subfigure}[t]{0.15\textwidth}
      \centering
       \includegraphics[width=\textwidth]{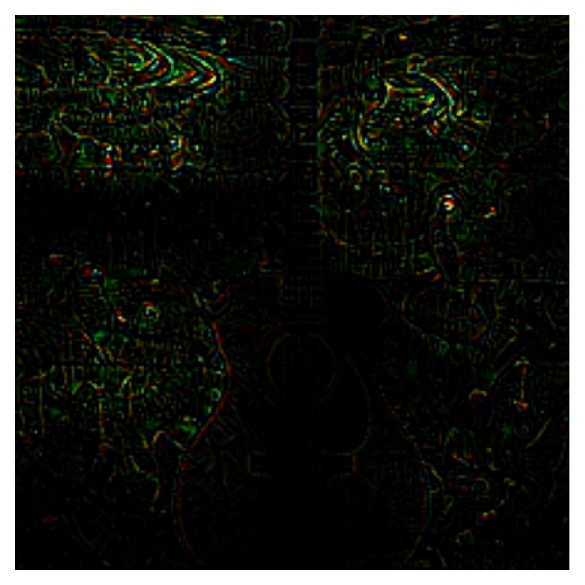}
         \vspace{-1em}
      \caption{Guided Grad-CAM 
      }
      \label{fig:method-det-grad-cam-adv}
  \end{subfigure}
  \centering
  \begin{subfigure}[t]{0.15\textwidth}
     \centering
      \includegraphics[width=\textwidth]{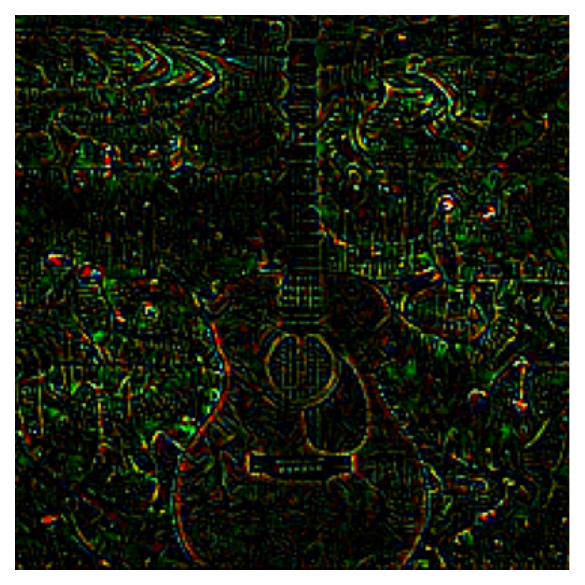}
        \vspace{-1em}
     \caption{ Guided Backprop 
     }
     \label{fig:method-det-guided-adv}
  \end{subfigure}
\vspace{-1.0 em}
  \caption{
    %
    Visualizations on the clean image~(Top) and the adversarial example (Bottom).
    \textbf{Left:} Original clean image and adversarial example. 
    \textbf{Middle:} Guided Grad-CAM visualization.
    \textbf{Right:} Guided Backpropagation visualization.
    %
  }
  \label{fig:method-cam}
  \vspace{-1.5 em}
\end{figure}

\begin{figure*}[!t]
  \centering
  \captionsetup[subfigure]{justification=centering}
  \begin{subfigure}[t]{0.48\textwidth}
      \centering
       \includegraphics[width=0.8\textwidth]{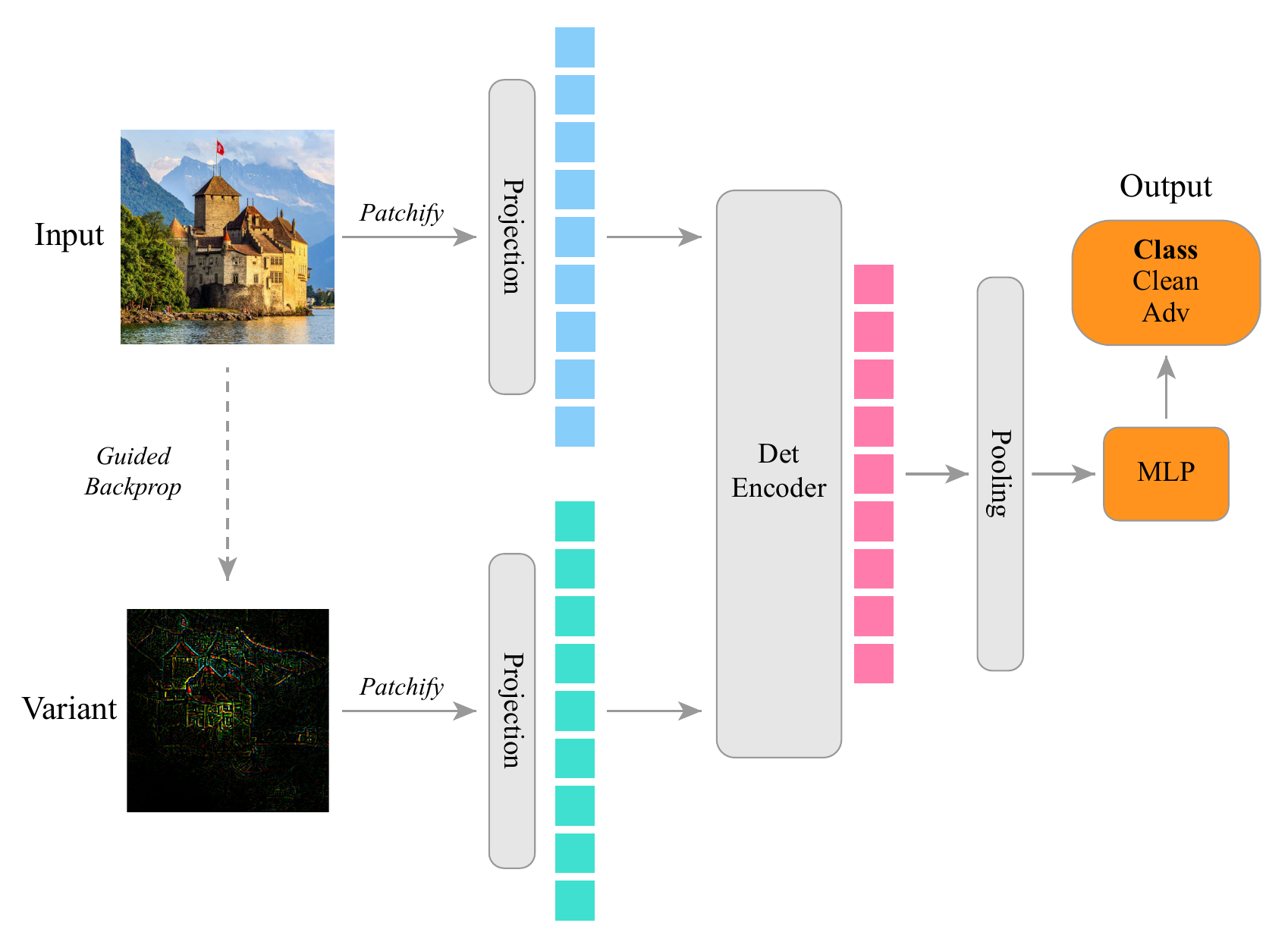}
      \caption{Detector}
      \label{fig:method-arch-det-attn}
  \end{subfigure}
\centering
\captionsetup[subfigure]{justification=centering}
\begin{subfigure}[t]{0.47\textwidth}
    \centering
     \includegraphics[width=\textwidth]{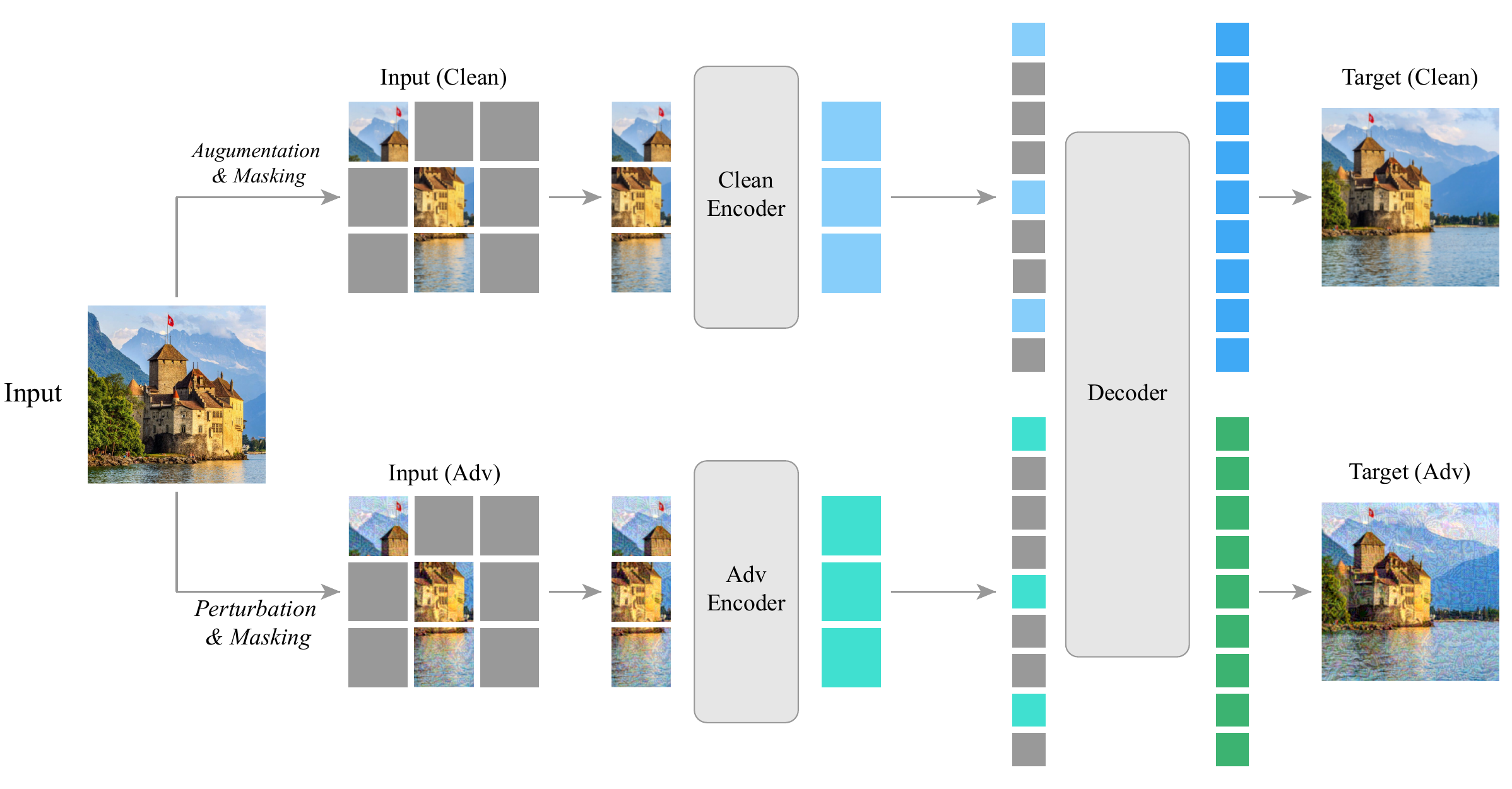}
    \caption{Classifer}
    \label{fig:method-arch-pre-train}
\end{subfigure}
\vspace{-0.5em}
\caption{
  Our model architecture: (a) detector and (b) classifier during the pre-training stage.  
}
\label{fig:method-arch-cls}
\vspace{-1.5 em}
\end{figure*}

The design of our detector architecture is motivated by our empirical observation in that 
\textit{the adversarial perturbation is detectable after Guided Backpropagation visualization.}
Due to the small distance between a clean image and its corresponding adversarial example,
their difference is notoriously imperceptible (see Figures~\ref{fig:method-det-cln} and \ref{fig:method-det-adv}),
making it theoretically hard to detect adversarial examples~\cite{tramer:icml22:hard-detect}.
In our empirical study, we resort to Guided Backpropagation~\cite{springenberg:iclrw15:guided} 
to visualize the difference between a clean image and an adversarial example.
Interestingly, we have discovered that after Guided Backpropagation visualization on the adversarial example, 
its adversarial perturbation is quite noticeable;
see Figure~\ref{fig:method-det-guided-cln} versus Figure~\ref{fig:method-det-guided-adv},
\ie\ visualization on a clean image versus on its adversarial example.
Notably, our experiments also include the visualization comparison of Guided Grad-CAM~\cite{selvaraju:iccv17:grad-cam}, developed recently;
see Figure~\ref{fig:method-det-grad-cam-cln} versus Figure~~\ref{fig:method-det-grad-cam-adv}.
However, Guided Grad-CAM exhibits inferior performance (compared to Guided Backpropagation)
in terms of exposing adversarial perturbation.
This empirical study motivates us to maximize the difference between clean images and adversarial examples 
by using Guided Backpropagation visualization.

Figure~\ref{fig:method-arch-det-attn} illustrates our detector architecture.
Given an input image $\bm{x} \in \mathbb{R}^{d}$,
we perform Guided Backpropagation 
on the original image,
arriving at an input variant $\bm{x}^{\prime} \in \mathbb{R}^{d}$.
%
%
Following the standard Vision Transformers~(ViT)~\cite{dosovitskiy:iclr21:vit},
we patchify the two inputs into two sets of image patches
and embed them via linear projection,
arriving at two sets of patch embeddings,
\ie\ $\mathbf{E}_{p} \in \mathbb{R}^{M \times D}$ and $\mathbf{E}_{p}^{\prime} \in \mathbb{R}^{M \times D} $, 
respectively for the original input and its input variant.
Here, $M$ represents the number of patches and $D$ indicates the hidden dimension.
Driven by the above empirical observation, 
a naive idea to expose adversarial perturbation is 
to add two sets of patch embeddings.
However, our empirical results 
show that this simple solution cannot achieve satisfactory performance. 
To address this issue, we propose a novel Multi-head Self-Attention (MSA)~\cite{vaswani2017attention}
to consider two sets of patch embeddings simultaneously,
inspired by recent studies~\cite{bao:icml20:UniLMv2,raffel:jmlr20:exploring,liu:iccv21:swin}.
Let $\mathbf{E} = \mathbf{E}_{p} +  \mathbf{E}_{\textrm{pos}} $ and $\mathbf{E}^{\prime} = \mathbf{E}_{p}^{\prime} +  \mathbf{E}_{\textrm{pos}} $
respectively represent two sets of patch embeddings after adding positional embeddings $\mathbf{E}_{\textrm{pos}} \in \mathbb{R}^{M \times D}$,
our proposed MSA can be expressed as follows:
\begin{equation} \label{eq:det-attn}
  \begin{gathered}
    \textrm{MSA}~( \bm{Q}, \bm{K}, \bm{V}) 
    = \textrm{Softmax} ( \frac{ \bm{QK}^{T} + \bm{B}}{\sqrt{d}} ) \bm{V}, \\
    \bm{Q} = \bm{W}_{Q} \cdot \bm{E}, ~~\bm{K} = \bm{W}_{K} \cdot \bm{E}, ~~\bm{V} = \bm{W}_{V} \cdot \bm{E}.
  \end{gathered}
\end{equation}
Here, $\bm{B} =\bm{W}_{B} \cdot \bm{E^{\prime}}$ is the relative detection bias 
obtained from the Guided Backpropagation-based input variant.
$\bm{W}_{Q}$, $\bm{W}_{K}$, $\bm{W}_{V}$, and $\bm{W}_{B}$ are learnable projection matrices,
similar to those in prior studies~~\cite{vaswani2017attention,rombach:cvpr22:stable_diffision}.
The intuition underlying \eqref{eq:det-attn} is that we aim to expose adversarial perturbation
by adding the relative bias obtained from Guided Backpropagation visualization.
After encoding, we follow Masked Autoencoders (MAE)~\cite{he:cvpr22:mae}
by performing global average pooling on the full set of encoded patch embeddings,
with the resulting token fed into an MLP (\ie\ multiple-layer perceptron) for telling whether the input is a clean image or not.

Aiming to further differentiate adversarial examples from clean images,
we propose a novel loss function to train our detector,
including a Cross-Entropy (CE) Loss $\mathcal{L}_{\rm{ce}}$
and a Soft-Nearest Neighbors (SNN) loss $\mathcal{L}_{\rm{snn}}$~\cite{salakhutdinov:aistats17:snn,frosst:icml19:snn},
for jointly penalizing the detection error and the similarity level between  the clean image
and the adversarial example, \ie\
\begin{equation} \label{eq:det-loss}
  \mathcal{L}_{\rm{det}} = (1 - \lambda) \cdot \mathcal{L}_{\rm{ce}} (g_{\bm{\phi}} (\bm{x}), y^{\det}) 
  + \lambda \cdot \mathcal{L}_{\rm{snn}} (\bm{z}^{\rm{cln}}, \bm{z}^{\rm{adv}}),
\end{equation}
where $\lambda \in (0, 1)$ is a hyperparameter to control the penalty degree of the two terms, 
and $\bm{z}^{\rm{cln}}$ and $\bm{z}^{\rm{adv}}$ denote the global representations, 
\ie\ the global average pooling of encoded representations,
for clean images and adversarial examples, respectively.

The SNN loss is a variant of contrastive loss, allowing for the inclusion of multiple positive pairs.
We regard members belonging to the same determined class (\eg\ two clean images) as positive pairs, 
%
while members belonging to different determined classes (\eg\ a clean image and an adversarial example) as negative pairs.
Given a mini-batch of $2B$ samples, with one half being clean images,
\ie\ $\{ (\bm{x}_{i}, ~y_{i}^{\rm{det}}$=$1) \}_{i=1}^{B}$,
and the other half of adversarial examples, 
\ie\ $\{ (\bm{x}_{i}^{\rm{adv}}, ~y_{i}^{\rm{det}}$=$-1) \}_{i=B+1}^{2B}$,
the SNN loss at temperature $\tau$ is defined below:
%
%
\begin{equation} \label{eq:loss-snn}
  \mathcal{L}_{\rm{snn}} = - \frac{1}{2B} \sum_{i=1}^{2B} \log 
  \frac{\sum_{j=1, i \neq j, y_{i}^{\rm{det}} = y_{j}^{\rm{det}}}^{2B} \exp (- \text{sim}(\bm{z}_{i}, \bm{z}_{j}) / \tau)  }
  {\sum_{j=1, i \neq k}^{2B} \exp (- \text{sim}(\bm{z}_{i}, \bm{z}_{k}) / \tau)},
\end{equation}
where $\bm{z}_{i}$ is the visual representations for the input $\bm{x}_{i}$ and the similarity metric $\text{sim}(\cdot, \cdot)$ is measured by the cosine distance.
The SNN loss enforces each point
to be closer to its positive pairs than to its negative pairs.
In other words, the SNN loss penalizes the similarity level between clean images and adversarial examples,
making adversarial examples more discernible by our detector.

\subsection{Classifer} 
\label{sec:cls}

Inspired by self-supervised learning for vision tasks~\cite{chen:icml20:sim-clr,bao:iclr22:beit,he:cvpr22:mae},
we separate our adversarial training into two stages, \ie\ pre-training and fine-tuning,
for learning high-quality visual representations and fine-tuning a robust classifier, respectively.

\begin{figure*}[!t]
  \centering
  \includegraphics[width=.85\textwidth]{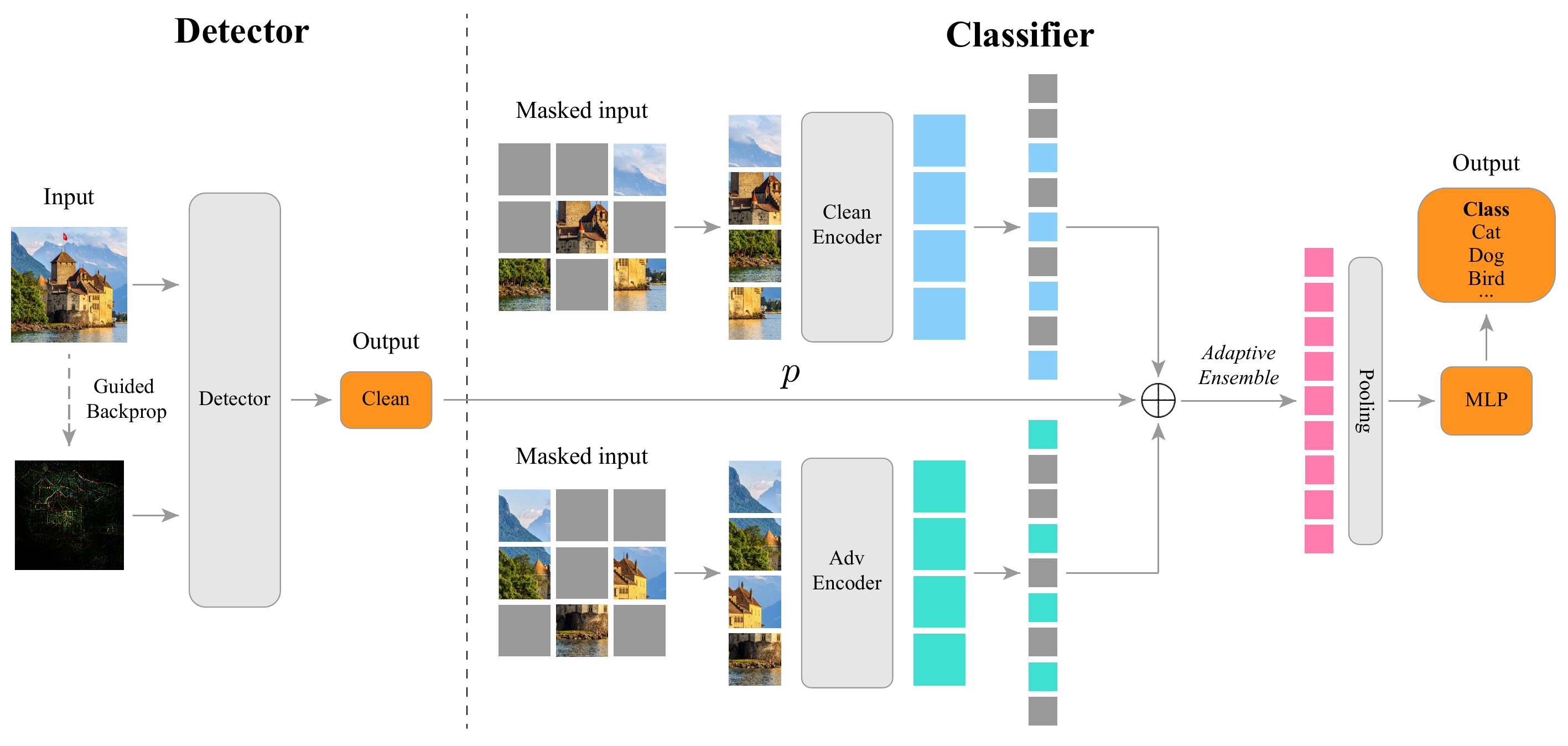}
    \vspace{-0.5 em}
  \caption{
    Illustration of our model architecture during the fine-tuning stage.
    } 
  \label{fig:method-arch}
  \vspace{-1.0 em}
\end{figure*}

\par\smallskip\noindent
{\bf Pre-training.}
Our classifier architecture for the pre-training is inspired by MAE~\cite{he:cvpr22:mae}.
Different from MAE, we utilize two encoders, denoted as the clean encoder and the adversarial encoder,
for learning visual representations from clean images and adversarial examples, respectively.
The decoder aims to reconstruct the original inputs
from the visual representations encoded by the two encoders.
Figure~\ref{fig:method-arch-pre-train} shows the classifier architecture during the pre-training.
Given an input image $ \bm{x} \in \mathbb{R}^{d}$,
let $\bm{x}^{\textrm{cln}}$ and $\bm{x}^{\textrm{adv}}$ denote its clean and adversarial variants, respectively, 
with the clean variant obtained by augmenting the original input.
%
Regarding the clean variant $\bm{x}^{\textrm{cln}}$,
we randomly mask out a large proportion of image patches (\eg\ $75 \%$)
and then feed the subset of visible patches into the clean encoder.
The masked tokens are inserted into corresponding positions after the encoder.
Finally, the decoder reconstructs the clean variant $\bar{\bm{x}}^{\text{cln}}$
from the full set of image patches,
including encoded visible patches and masked tokens.
The reconstruction of the adversarial variant $\bm{x}^{\textrm{adv}}$ follows a similar procedure,
except that its visible patches are encoded by the adversarial encoder.
Notably, the position of masked image patches in the adversarial variant $\bm{x}^{\textrm{adv}}$ 
is the same as that in the clean variant $\bm{x}^{\textrm{cln}}$
in order to minimize their visual representation difference
during the pre-training.

Let $\bar{\bm{z}}^{\textrm{cln}}$  and $\bar{\bm{z}}^{\textrm{adv}}$ respectively 
denote the global representations of clean and adversarial variants,
obtained by performing global average pooling on the decoder's input sequence.
%
Our design utilizes a new loss function to learn visual representations
by simultaneously minimizing the reconstruction error and the visual representation difference, \ie\
\begin{equation} \label{eq:loss-enc}
  \mathcal{L}_{\textrm{enc}} = (1 - \Omega) \cdot \mathcal{L}_{\rm{rec}} (\bm{x}, \bar{\bm{x}}) 
  + \Omega \cdot \mathcal{L}_{\rm{cl}} (\bar{\bm{z}}^{\rm{cln}}, \bar{\bm{z}}^{\rm{adv}} ),
\end{equation}
where $\Omega \in (0,1)$ is a hyperparameter and $\bar{\bm{x}}$ is the reconstructed image. 
$\mathcal{L}_{\textrm{rec}}$ and $\mathcal{L}_{\textrm{cl}}$
denote the reconstruction loss and the contrastive loss, respectively.
Given a set of $B$ input images, we first generate their adversarial variants,
arriving at a mini-batch of $2B$ samples, consisting of $B$ clean variants $\{\bm{x}_{i}^{\textrm{cln}} \}_{i=1}^{B}$
and $B$ adversarial variants $\{\bm{x}_{i}^{\rm{adv}} \}_{i=B+1}^{2B}$.
We consider the form of contrastive loss in SimCLR~\cite{chen:icml20:sim-clr},  
and define our contrastive loss at temperature $\tau$ as follows:
\begin{equation} \label{eq:loss-sim-clr}
  \begin{aligned}
    \ell (i,j) &= - \log \frac{\exp(\text{sim}(\bar{\mathbf{z}}_{i}, \bar{\mathbf{z}}_{j}) / \tau )}
    {\sum_{i \neq k, k = 1, \dots, 2B} \exp (\text{sim}(\bar{\mathbf{z}}_{i}, \bar{\mathbf{z}}_{k}) / \tau  ) }, \\
    & \mathcal{L}_{\rm{cl}} = \frac{1}{2B} \sum_{k=1}^{B} \left[ \ell(k, k+B) +\ell(k+B, k) \right],  
  \end{aligned}  
\end{equation}
where $\bar{\bm{z}}_{i}$ denotes visual representations for $\bm{x}^{\textrm{cln}}_{i}$ (or $\bm{x}^{\textrm{adv}}_{i}$) and the similarity level $\text{sim} (\cdot, \cdot)$ is measured by the cosine distance. 
In particular, we regard clean and adversarial variants from the same input as  positive pairs,
while the rest in the same batch are negative pairs.
Hence, the loss value decreases when visual representations 
for the clean and the adversarial variants of the same input become more similar.

\par\smallskip\noindent
{\bf Fine-tuning.}
%
The detector and the classifier (including two encoders and one decoder) are trained jointly in the pre-training stage.
After that, we drop the decoder and freeze the weights in the well-trained detector and two encoders,
with Figure~\ref{fig:method-arch} depicting our model architecture 
during the fine-tuning stage.
Different from MAE, 
which encodes the full set of image patches during the fine-tuning,
our approach randomly masks out a relatively small proportion of image patches (\eg\ $45\%$),
aiming to eliminate the potential adversarial effect if the input is an adversarial example.

Given an input image $(\bm{x}, y^{\text{cls}})$,
where $\bm{x} \in \mathbb{R}^{d}$ 
is either a clean image or an adversarial example with the label $y^{\text{cls}} \in [C]$,
we randomly mask the input image twice, arriving at two different masked inputs.
Two subsets of visible patches from the two masked inputs are fed into the clean and the adversarial encoders, respectively.
The masked tokens are introduced onto their corresponding positions after the encoder,
obtaining two full sets of visual representations,
\ie\ $\hat{\bm{z}}^{\text{cln}}$ and $\hat{\bm{z}}^{\text{adv}}$ which are partially encoded by the clean and the adversarial encoders,
respectively.
We then perform the global average pooling on the \textit{adaptive ensemble} of $\hat{\bm{z}}^{\text{cln}}$ and $\hat{\bm{z}}^{\text{adv}}$,
with the result fed into an MLP for classification.
\par\smallskip\noindent
{\bf Adaptive Ensemble.}
Although randomly masking an input image can eliminate the potential adversarial effect,
this way inevitably hurts standard accuracy during the fine-tuning.
In this paper, we propose \textit{adaptive ensemble}~\cite{li2023ensemble} to tackle this issue.
That is, the global representation for an input image is derived from 
the sum of $\hat{\bm{z}}^{\text{cln}}$ and $\hat{\bm{z}}^{\text{adv}}$
with an adaptive factor $p \in [0, 1]$,
where $\hat{\bm{z}}^{\text{cln}}$ and $\hat{\bm{z}}^{\text{adv}}$
are visual representations encoded by the clean and the adversarial encoders, respectively,
and $p$ is the probability of the input image 
being a clean image estimated by our detector.

Let $A$ be a full set of image patches
and $V$ be a subset of $A$, including visible patches only.
$\mathbb{1}_{V} (\cdot)$ is the indicator function for evaluating whether an image patch is visible.
Hence, for each image patch of $A$, we have,
\begin{equation} \label{eq:vis-indicator}
  \mathbb{1}_{V}(i) = \begin{cases}
    1, ~\text{if the patch is visible} \\
    0, ~\text{otherwise}
  \end{cases},
  ~i = 1,2, \small{\dots}, M,
\end{equation} 
where $M$ is the number of image patches, \ie\ $|A|$.
For notational convenience, we let $\mathbb{1}_{V}^{\text{cln}}  $ 
indicate visible patches fed into the clean encoder. 
Likewise, $\mathbb{1}_{V}^{\text{adv}}$ 
indicates visible patches fed into the adversarial encoder.  
Let $\hat{\bm{z}}_{i}$ be the visual representation of the $i$-th image patch,
with $i \in \{1,2, \dots, M \}$.
%
Our adaptive ensemble is defined by:
\begin{equation} \label{eq:adapt-ensem}
    \hat{\bm{z}}_{i} = \frac{p \cdot \mathbb{1}_{V}^{\text{cln}} (i) \cdot \hat{\bm{z}}_{i}^{\text{cln}} 
    + (1-p) \cdot \mathbb{1}_{V}^{\text{adv}}(i) \cdot \hat{\bm{z}}_{i}^{\text{adv}} }
    { \max \left( p \cdot \mathbb{1}_{V}^{\text{cln}} (i) + (1-p) \cdot \mathbb{1}_{V}^{\text{adv}} (i), ~\epsilon \right) },
\end{equation}
where the denominator serves to normalize the adaptive ensemble of  $\hat{\bm{z}}^{\text{cln}}_{i}$ and $\hat{\bm{z}}^{\text{adv}}_{i}$, 
and $\epsilon$ is a small value to avoid divison by zero (\ie\, $\epsilon = 1e-12$ in this paper).
The intuition underlying Eq.~(\ref{eq:adapt-ensem}) is that if our detector has a high confidence
that the input is a clean image (\ie\ $p$ is large), 
the global representation $\hat{\bm{z}}_{i}$
will be mostly encoded by the clean encoder.
Otherwise, $\hat{\bm{z}}_{i}$ will be mainly encoded by the adversarial encoder.
In addition, as our pre-training encourages the similarity level of
the clean and the adversarial variants from a given input
(see Eq.~(\ref{eq:loss-enc}) and Eq.~(\ref{eq:loss-sim-clr})),
and two different masked inputs exist upon the fine-tuning,
the invisible image patches in one masked input can be glimpsed from the other masked input.

\section{Experiments and Results}\label{sec:exp}

%
\begin{table*} [htbp] 
    \small
    \centering
    \setlength\tabcolsep{12 pt}
   \caption{Model Details used in our design
   }
   \vspace{-0.5 em}
   \begin{tabular}{@{}ccccccc@{}}
    \toprule
    \multicolumn{2}{c}{Model}                                       & Layer & Hidden Size & Head & MLP Size & Parameters \\ \midrule
    \multicolumn{1}{c}{Detector}                    & ViT-Tiny     & 12    & 192         & 3    & 768      & 7.8M       \\ \midrule
    \multicolumn{1}{c}{\multirow{3}{*}{Classifier}} & Clean Encoder & 12    & 384         & 3    & 1536     & 21.3M      \\ 
    \multicolumn{1}{c}{}                            & Adv Encoder   & 12    & 384         & 3    & 1536     & 21.3M      \\ 
    \multicolumn{1}{c}{}                            & Decoder       & 8     & 192         & 4    & 768      & 3.6M       \\ \bottomrule
    \end{tabular}
   \label{tab:exp-model-size}
\end{table*}

\subsection{Experimental Setup}
\label{sec:exp-setup}

\par\smallskip\noindent
{\bf Datasets.}
We conduct experiments on three widely-used benchmarks.
(\romsm{1}) \textbf{CIFAR-10}~\cite{krizhevsky:09:cifar}: 
$60,000$ $32\text{x}32$ RGB images of $10$ classes.
(\romsm{2}) \textbf{CIFAR-100}~\cite{krizhevsky:09:cifar}: 
$60,000$ $32\text{x}32$ RGB examples in $100$ categories.
(\romsm{3}) \textbf{Tiny-ImageNet}~\cite{deng:cvpr09:imagenet}: 
$120,000$ $64\text{x}64$ RGB images of $200$ classes.

\par\smallskip\noindent
{\bf Compared Methods.}
We compare our approach with four detection methods,
\ie\ \textbf{Odds}~\cite{roth:icml19:odds}, \textbf{NIC}~\cite{ma:ndss19:nic}, 
\textbf{GAT}~\cite{yin:iclr20:gat},
and~\textbf{JTLA}~\cite{raghuram:icml21:jtla}.
%
%
We compare our approach with five adversarial training (AT) counterparts:
\textbf{PGD-AT}~\cite{madry:iclr18:pgd}, \textbf{TRADES}~\cite{zhang:icml19:trades},
\textbf{FAT}~\cite{wong:iclr20:fast}, \textbf{Sub-AT}~\cite{li:cvpr22:sub-at}, and \textbf{LAS-AWP}~\cite{jia:cvpr22:las},
to exhibit how it boosts the ViT's robustness.
%

%
\par\smallskip\noindent
{\bf Evaluation.}
We consider three \sota\ adaptive attacks, \ie\ \textbf{AutoAttack}~\cite{croce:icml20:autoattack}, 
\textbf{Adaptive Auto Attack ($\text{A}^{3}$)}~\cite{yao:nips21:a3},
and \textbf{Parameter-Free Adaptive Auto Attack ($\text{PF-A}^{3}$)}~\cite{liu:cvpr22:pf_a3},
for evaluating detection accuracy and adversarial robustness.
The attack constraint, if not specified, is set to $\epsilon = 8/255$.

\par\smallskip\noindent
{\bf Model Size.}
We build our detector and classifier on top of Vision Transformers (ViT),
with their architectures following 
ViT~\cite{dosovitskiy:iclr21:vit}
and MAE~\cite{he:cvpr22:mae}, respectively.
Our model size is pruned down to as small as possible in order to conduct a fair comparison with baselines.
Table~\ref{tab:exp-model-size} lists the model size details.
Our architecture consists of a detector and a classifier (including two encoders and one decoder),
with $54.0M$ parameters in total.
To conduct a fair comparison,
existing adversarial training baselines use 
the ViT-Base model~\cite{dosovitskiy:iclr21:vit}
with total parameters of $85.6M$ as the backbone network.

\par\smallskip\noindent
{\bf Hyperparameters.}
For all our models, if not specified, we use AdamW~\cite{loshchilov:iclr19:adamw}
with $\beta_{1}$=$0.9$, $\beta_{2}$=$0.999$, the weight decay of $0.05$, and a batch size of $512$.
During the pre-training, the detector and the classifier (\ie\ two encoders and one decoder) are trained jointly.
For the detector, we follow the setting in~\cite{goyal:2017:base_lr}
by setting the epochs of $100$, the base learning rate of $1e-3$, the linear warmup epochs of $5$,
and the cosine decay schedule~\cite{loshchilov:iclr17:cosine}.
For the classifier, by contrast, we pre-train it for $200$ epochs, with the base learning rate of $1e-4$,
the linear warmup of $20$ epochs, and a masking ratio of $75\%$.
After pre-training, we drop the decoder and freeze the weights on the detector and the two encoders.
Then, we finetune the classifier
for $100$ epochs, with the base learning rate of $1e-3$, 
the linear warmup of $5$, and the cosine decay schedule,
and a masking ratio of $45\%$.
The patch size is set to $4$ (or $8$) for CIFAR-10/CIFAR-100 (or Tiny-ImageNet).
We grid-search hyperparamters $\lambda$ in \eqref{eq:det-loss} and $\Omega$ in \eqref{eq:loss-enc} 
of Section~\ref{sec:our_approach}
and empirically set $\lambda$ to $0.15$ and  $\Omega$ to $0.35$ for all datasets.


%

\begin{figure*} [!t] 
  \centering
  \captionsetup[subfigure]{justification=centering}
  \begin{subfigure}[t]{0.24\textwidth}
      \centering
       \includegraphics[width=\textwidth]{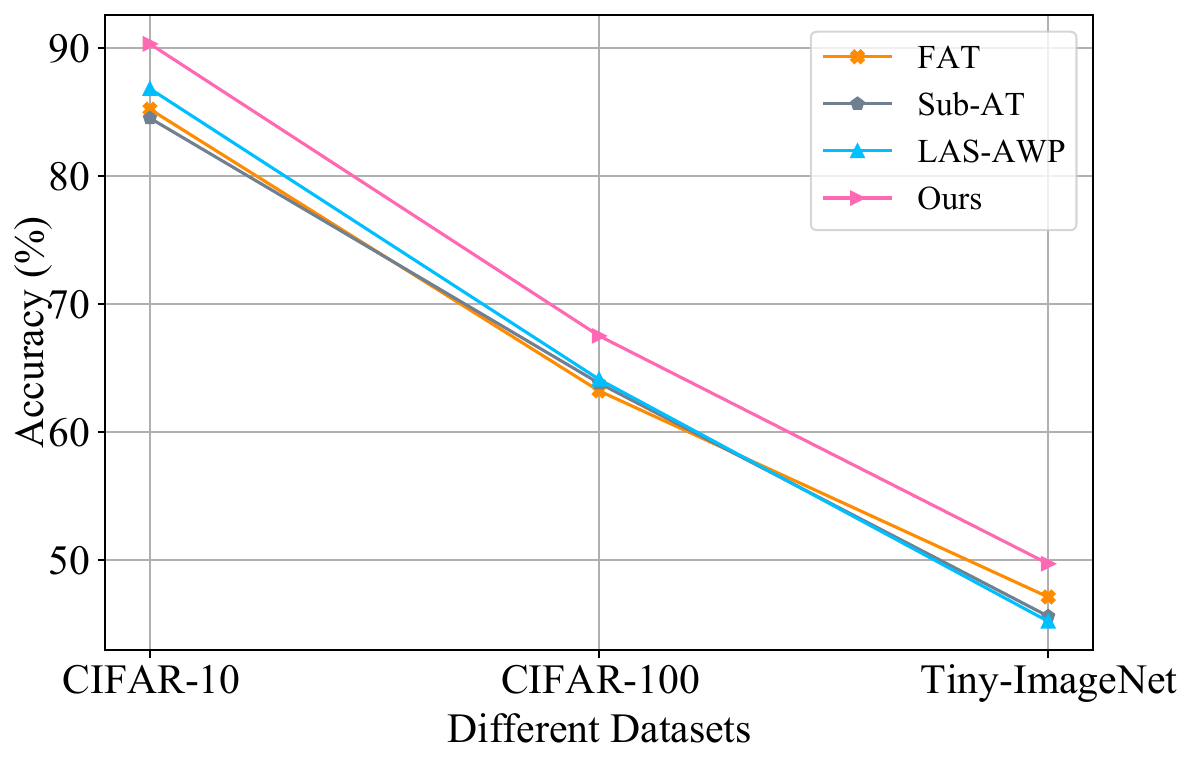}
         \vspace{-1.5em}
      \caption{ Standard Accuracy
      }
      \label{fig:exp-stable-clean}
  \end{subfigure}
  \centering
   \begin{subfigure}[t]{0.24\textwidth}
      \centering
       \includegraphics[width=\textwidth]{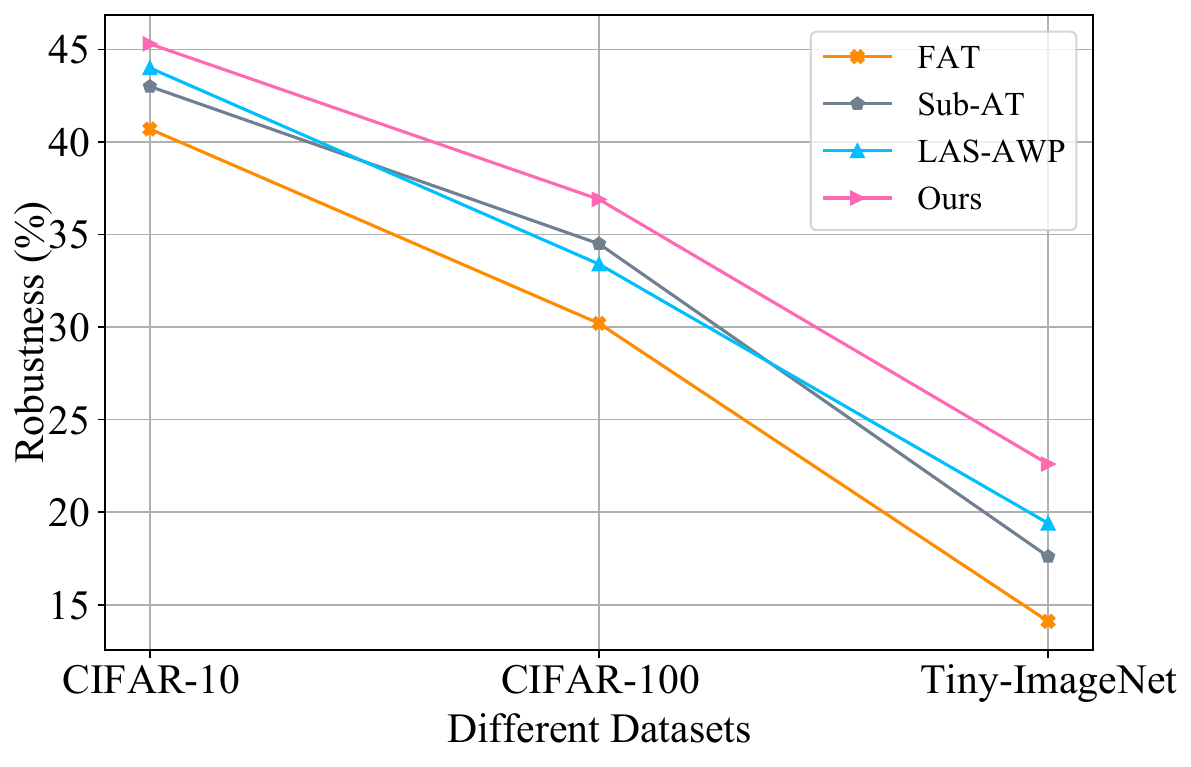}
         \vspace{-1.5em}
      \caption{Under AutoAttack
      }
      \label{fig:exp-stable-adv-aa}
  \end{subfigure}
  \captionsetup[subfigure]{justification=centering}
  \begin{subfigure}[t]{0.24\textwidth}
      \centering
       \includegraphics[width=\textwidth]{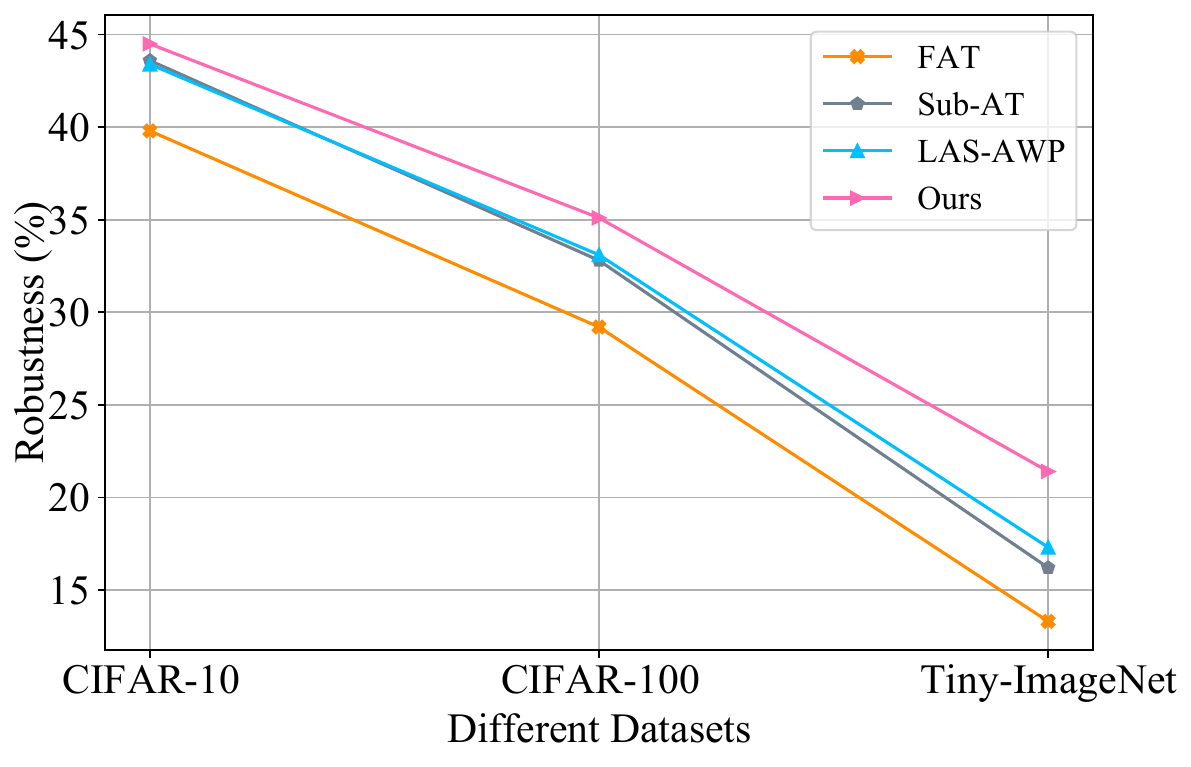}
         \vspace{-1.5em}
      \caption{ Under A$^{3}$ attack
      }
      \label{fig:exp-stable-adv-a3}
  \end{subfigure}
  \captionsetup[subfigure]{justification=centering}
  \begin{subfigure}[t]{0.24\textwidth}
      \centering
       \includegraphics[width=\textwidth]{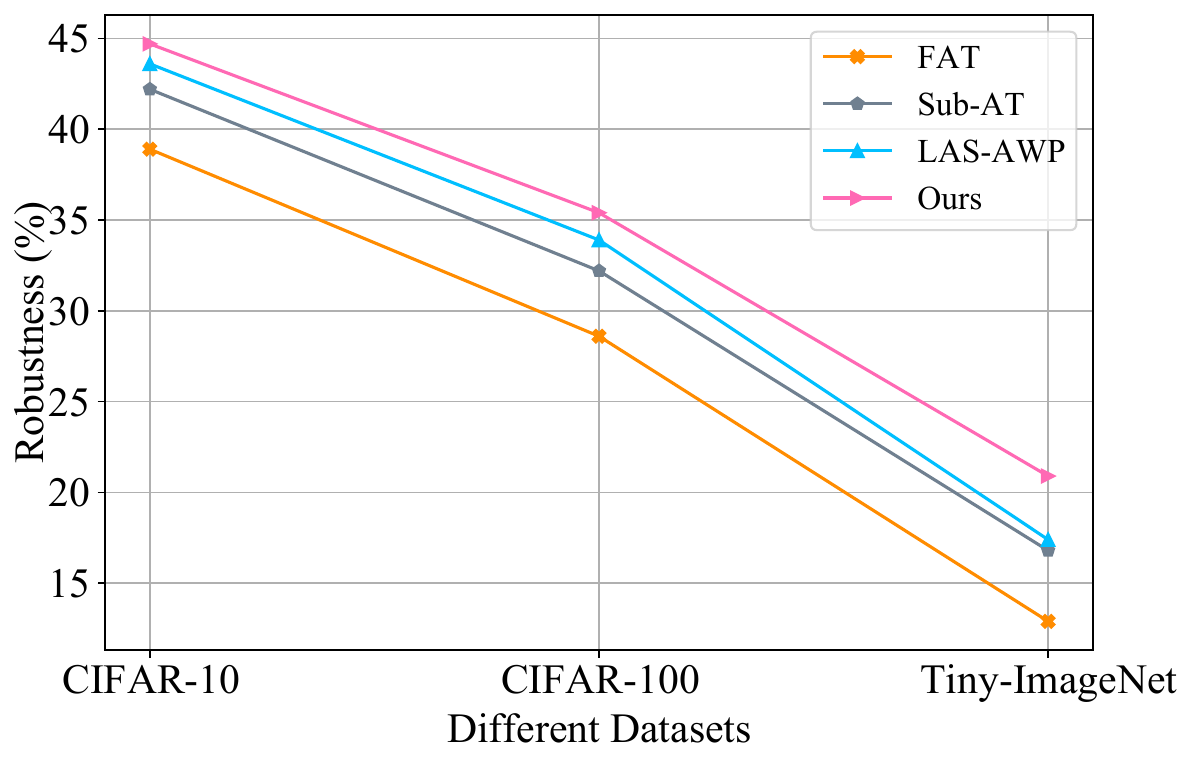}
         \vspace{-1.5em}
      \caption{Under PF-A$^{3}$ attack
      }
      \label{fig:exp-stable-adv-pf-a3}
  \end{subfigure}
\vspace{-0.5 em}
  \caption{
      Performance stability under different datasets 
      and different adaptive attacks.
  }
  \label{fig:exp-stale}
  \vspace{-1.0 em}
\end{figure*}

\subsection{Overall Performance on Our Classifier}
\label{sec:exp-cls}

\begin{table} [!t] 
   \small
   \centering
   \setlength\tabcolsep{2 pt}
  \caption{Overall comparative results of standard accuracy and adversarial robustness on CIFAR-10
  under attack constraints of $\epsilon= 4/255$ and of $\epsilon= 8/255$,
  with best results shown in bold
  }
  \vspace{-0.5 em}
  \begin{tabular}{@{}c|c|ccc|ccc@{}}
   \toprule
   \multirow{2}{*}{Method} &
     \multirow{2}{*}{\begin{tabular}[c]{@{}c@{}}Standard \\ Accuracy\end{tabular}} &
     \multicolumn{3}{c|}{ Robustness ( $\epsilon$ = $4 / 255 $) } &
     \multicolumn{3}{c}{ Robustness ( $\epsilon$ = $8 / 255 $) } \\ \cmidrule(l){3-8} 
           &      & \multicolumn{1}{c}{AutoAttack} & \multicolumn{1}{c}{A$^{3}$}   & PF-A$^{3}$ & \multicolumn{1}{c}{AutoAttack} & \multicolumn{1}{c}{A$^3$}   & PF-A$^3$ \\ \midrule
   PGD-AT  & 83.7 & \multicolumn{1}{c}{45.1}       & \multicolumn{1}{c}{43.4} & 43.5  & \multicolumn{1}{c}{41.5}       & \multicolumn{1}{c}{40.3} & 40.9  \\ 
   TRADES  & 84.9 & \multicolumn{1}{c}{46.6}       & \multicolumn{1}{c}{43.5} & 43.6  & \multicolumn{1}{c}{41.5}       & \multicolumn{1}{c}{40.7} & 40.3  \\ 
   FAT     & 85.2 & \multicolumn{1}{c}{43.8}       & \multicolumn{1}{c}{42.1} & 43.4  & \multicolumn{1}{c}{40.7}       & \multicolumn{1}{c}{39.8} & 38.9  \\ 
   Sub-AT  & 84.5 & \multicolumn{1}{c}{47.4}       & \multicolumn{1}{c}{46.9} & 45.1  & \multicolumn{1}{c}{43.3}       & \multicolumn{1}{c}{43.6} & 42.2  \\ 
   LAS-AWP & 86.8 & \multicolumn{1}{c}{46.9}       & \multicolumn{1}{c}{45.0} & 47.7  & \multicolumn{1}{c}{44.0}       & \multicolumn{1}{c}{43.4} & 43.6  \\ \midrule
   Ours &
     \textbf{90.3} &
     \multicolumn{1}{c}{\textbf{49.8}} &
     \multicolumn{1}{c}{\textbf{49.5}} &
     \textbf{48.1} &
     \multicolumn{1}{c}{\textbf{45.3}} &
     \multicolumn{1}{c}{\textbf{44.5}} &
     \textbf{44.7} \\ \bottomrule
   \end{tabular}
  \label{tab:cls-cifar10}
  \vspace{-1.0 em}
\end{table}

\par\smallskip\noindent
{\bf Overall Comparisons on CIFAR-10.}
We first conduct extensive experiments on CIFAR-10
and compare our approach to its state-of-the-art adversarial training (AT) counterparts
listed in Section~\ref{sec:exp-setup} in terms of 
standard accuracy and adversarial robustness under attack constraints 
of $\epsilon= 4/255$ and of $\epsilon= 8/255$.
Table~\ref{tab:cls-cifar10} lists comparative results.
It is observed that our approach achieves the best performance under all three scenarios.
In particular, our approach achieves the standard accuracy of $90.3\%$,
outperforming the best competitor (\ie\ LAS-AWP) by $3.5\%$.
This is contributed by employing two encoders to extract visual representations respectively from clean images
and adversarial examples, able to significantly mitigate the adverse effect of adversarial training on standard accuracy.
Besides, when the attack constraint is set to $\epsilon= 4/ 255$,
our approach achieves the best robustness of $49.8 \%$, $49.5 \%$, and $48.1 \%$
against AutoAttack, Adaptive Auto Attack (A$^{3}$), 
and Parameter-Free Adaptive Auto Attack (PF-A$^{3}$), respectively.
Our method significantly surpasses all its counterparts.
For example, it outperforms two recent state-of-the-arts,
\ie\ Sub-AT and LAS-AWP, respectively by $2.6 \%$ and $4.5 \%$ 
under the attack of A$^{3}$.
Thirdly, increasing the attack constraint to $\epsilon= 8/255$ results in the decrease of adversarial robustness.
But our approach still maintains the best robustness of $45.3 \%$, $44.5 \%$, and $44.7 \%$
under the attack of AutoAttack, A$^{3}$, and PF-A$^{3}$, respectively.
The comparative results demonstrate that our masked adaptive ensemble
is robust enough to withstand strong white-box attacks.
This is because masking a small proportion of image patches 
can significantly mitigate the adversarial effect of malicious inputs.

%
\begin{table}  [!t]  
    \scriptsize
    \centering
    \setlength\tabcolsep{3 pt}
   \caption{Overall comparisons on CIFAR-100 and Tiny-ImageNet, with best results shown in bold
   }
   \vspace{-0.5 em}
   \begin{tabular}{@{}c|cccc|cccc@{}}
    \toprule
    \multirow{2}{*}{Method} &
      \multicolumn{4}{c|}{CIFAR-100} &
      \multicolumn{4}{c}{Tiny-ImageNet} \\ \cmidrule(l){2-9} 
     &
      \multicolumn{1}{c}{\begin{tabular}[c]{@{}c@{}}Standard \\ Accuracy\end{tabular}} &
      \multicolumn{1}{c}{AutoAttack} &
      \multicolumn{1}{c}{A$^3$} &
      PF-A$^{3}$ &
      \multicolumn{1}{c}{\begin{tabular}[c]{@{}c@{}}Standard \\ Accuracy\end{tabular}} &
      \multicolumn{1}{c}{AutoAttack} &
      \multicolumn{1}{c}{A$^3$} &
      PF-A$^{3}$ \\ \midrule
    PGD-AT &
      \multicolumn{1}{c}{62.5} &
      \multicolumn{1}{c}{31.9} &
      \multicolumn{1}{c}{31.5} &
      31.1 &
      \multicolumn{1}{c}{42.9} &
      \multicolumn{1}{c}{17.2} &
      \multicolumn{1}{c}{16.4} &
      16.8 \\ 
    TRADES &
      \multicolumn{1}{c}{61.8} &
      \multicolumn{1}{c}{32.5} &
      \multicolumn{1}{c}{31.3} &
      31.6 &
      \multicolumn{1}{c}{44.1} &
      \multicolumn{1}{c}{15.2} &
      \multicolumn{1}{c}{14.7} &
      14.1 \\ 
    FAT &
      \multicolumn{1}{c}{63.2} &
      \multicolumn{1}{c}{30.2} &
      \multicolumn{1}{c}{29.2} &
      28.6 &
      \multicolumn{1}{c}{47.1} &
      \multicolumn{1}{c}{14.1} &
      \multicolumn{1}{c}{13.3} &
      12.9 \\ 
    Sub-AT &
      \multicolumn{1}{c}{63.8} &
      \multicolumn{1}{c}{34.5} &
      \multicolumn{1}{c}{32.8} &
      32.2 &
      \multicolumn{1}{c}{45.6} &
      \multicolumn{1}{c}{17.6} &
      \multicolumn{1}{c}{16.2} &
      16.8 \\ 
    LAS-AWP &
      \multicolumn{1}{c}{64.1} &
      \multicolumn{1}{c}{33.4} &
      \multicolumn{1}{c}{33.1} &
      33.9 &
      \multicolumn{1}{c}{45.2} &
      \multicolumn{1}{c}{19.4} &
      \multicolumn{1}{c}{17.3} &
      17.4 \\ \midrule
    Ours &
      \multicolumn{1}{c}{\textbf{67.5}} &
      \multicolumn{1}{c}{\textbf{36.9}} &
      \multicolumn{1}{c}{\textbf{35.1}} &
      \textbf{35.4} &
      \multicolumn{1}{c}{\textbf{49.7}} &
      \multicolumn{1}{c}{\textbf{22.6}} &
      \multicolumn{1}{c}{\textbf{21.4}} &
      \textbf{20.9} \\ \bottomrule
    \end{tabular}
   \label{tab:cls-cifar100-tiny-imagenet}
   \vspace{-1.0 em}
\end{table}

\par\smallskip\noindent
{\bf Overall Comparisons on CIFAR-100 and Tiny-ImageNet.}
Here, we conduct a comprehensive comparison
between our approach and adversarial training (AT) counterparts
on CIFAR-100 and Tiny-ImageNet datasets.
Table~\ref{tab:cls-cifar100-tiny-imagenet} lists the comparative results.
On CIFAR-100, we observed that our approach achieves the best standard accuracy
of $67.5 \%$, outperforming the best competitor (\ie\ LAS-AWP) by $3.4 \%$.
Meanwhile, our method achieves the best robustness 
of $36.9 \%$, $35.1 \%$, and $35.4 \%$ under the attack 
of AutoAttack, A$^{3}$, and PF-A$^{3}$, respectively.
This confirms that our approach can achieve a decent standard accuracy and robustness
when being generalized to the dataset with large classes. 
On the Tiny-ImageNet dataset, both our approach and the baseline methods experience a decrease in performance. However, our proposed method still achieves the highest standard accuracy of $49.7 \%$, 
which outperforms the best baseline (\ie\ FAT), by $2.6 \%$.
Moreover, all baselines suffer from a poor robustness on the Tiny-ImageNet dataset
(\ie\ $\leq 20.0 \%$), while our approach maintains a decent robustness
of $22.6 \%$, $21.4 \%$, and $20.9 \%$ under the attack of AutoAttack, A$^{3}$, and PF-A$^{3}$, respectively.

\par\smallskip\noindent
{\bf Performance Stability.}
We next conduct experiments on CIFAR-10, CIFAR-100, and Tiny-ImageNet
to evaluate the performance stability 
under different scales of datasets and different types of adaptive attacks.
We compare our approach with three baselines,
\ie\ FAT, Sub-AT, and LAS-AWP.
Figures~\ref{fig:exp-stable-clean}, \ref{fig:exp-stable-adv-aa}, 
\ref{fig:exp-stable-adv-a3} and \ref{fig:exp-stable-adv-pf-a3}
illustrate the comparative results of standard accuracy, 
as well as robustness against AutoAttack, A$^{3}$, and PF-A$^{3}$,
respectively.
We have three discoveries.
First, as depicted in Figure~\ref{fig:exp-stable-clean},
our approach (\ie\ the pink line) achieves the best standard accuracies 
of $90.3 \%$, $67.5 \%$, and $49.7 \%$ under CIFAR-10, CIFAR-100,
and Tiny-ImageNet, respectively.
The empirical evidence verifies that our approach 
can maintain superior standard accuracy when generalized to large datasets.
Second, on all three datasets, our approach achieves the best robustness under all adaptive attacks,
as shown in Figures~\ref{fig:exp-stable-adv-aa}, 
\ref{fig:exp-stable-adv-a3} and \ref{fig:exp-stable-adv-pf-a3}.
Take the robustness results under PF-A$^{3}$ (\ie\ Figure~\ref{fig:exp-stable-adv-pf-a3}) for example,
our proposed masked adaptive ensemble achieves the robustness
of $44.7 \%$, $35.4 \%$, and $ 20.9 \%$ on CIFAR-10, CIFAR-100, and Tiny-ImageNet, respectively.
These results outperform those of LAS-AWP (\ie\ the blue line), which is the best baseline,
by $3.1 \%$, $4.1 \%$, and $3.5 \%$, respectively.
Third, when scaling up the dataset from CIFAR-10 to Tiny-ImageNet,
our approach suffers from the least robustness degradation 
of $22.7 \%$, $23.1 \%$, and $23.8 \%$ under the attack of AutoAttack, A$^{3}$, and PF-A$^{3}$, respectively.
These results confirm that in terms of robustness, our approach enjoys the best performance stability upon scaling up to large datasets.


\begin{figure*}[!t]
  \centering
  \includegraphics[width=.85\textwidth]{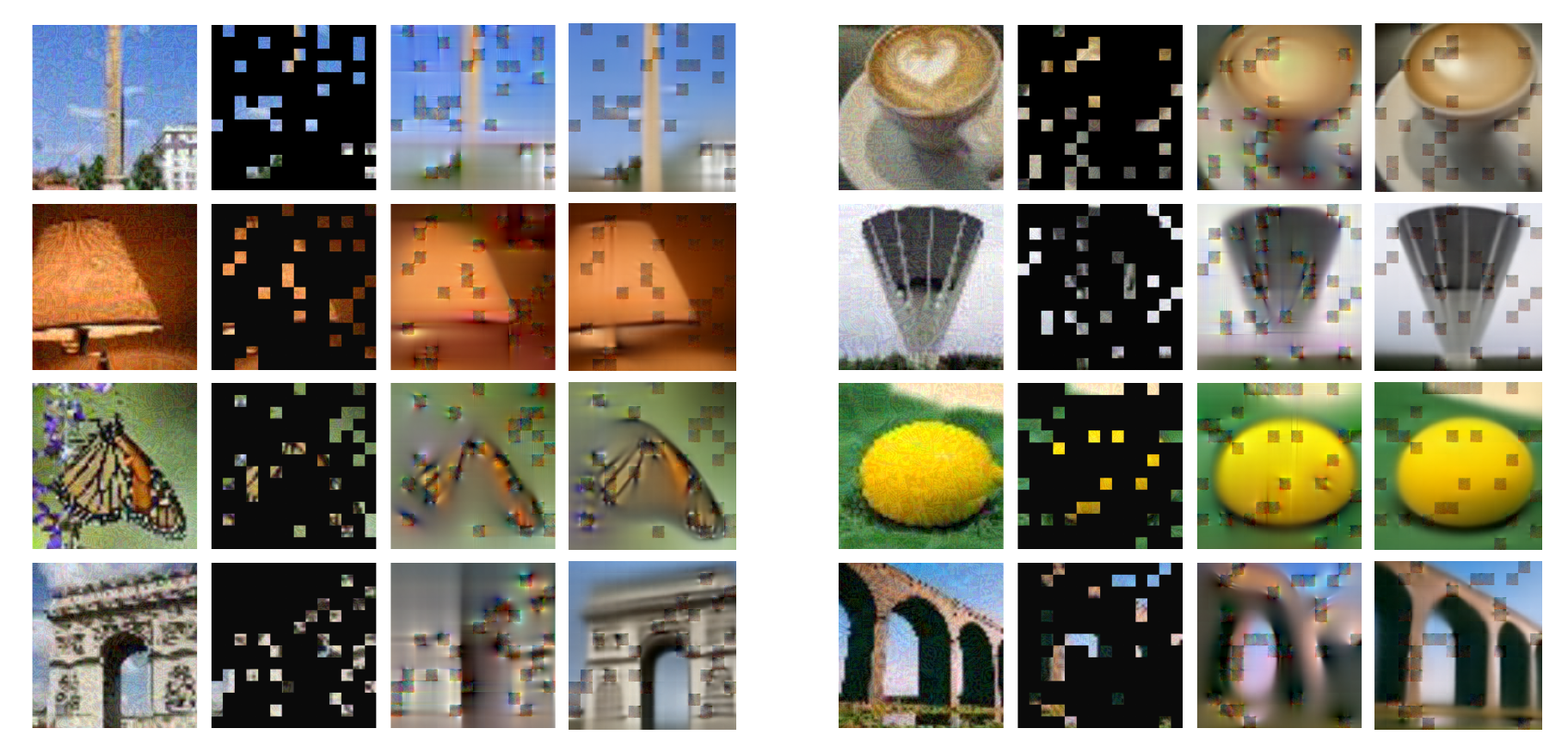}
  \vspace{-1.0 em}
  \caption{
    Comparison of the reconstruction quality from masked adversarial examples 
    by employing our approach with/without the contrastive loss,
    denoted as our approach (w/ CL) and our approach (w/o CL), respectively.
    From left to right are the original adversarial example, the masked input, reconstruction by our approach (w/o CL), and reconstruction by our approach (w/ CL), respectively.
    } 
  \label{fig:exp-rec}
  \vspace{-0.5 em}
\end{figure*}

\subsection{Ablation Studies on Our Classifier}\label{sec:as-cls}

\par\smallskip\noindent
{\bf Pre-training: Contrastive Loss.}
We qualitatively and quantitatively exhibit 
the impact of our proposed loss, \ie\ \eqref{eq:loss-sim-clr}, 
on learning visual representations.
We first present the qualitative evaluations.
Specifically, we reconstruct masked adversarial examples 
and compare reconstruction quality by utilizing our approach 
with/without the contrastive loss (CL) in SimCLR~\cite{chen:icml20:sim-clr}.
Figure~\ref{fig:exp-rec} illustrates the qualitative results.
For images on each row, from left to right,  are 
original adversarial example, the masked input,
the image generated by our approach without the CL (\ie\ w/o CL), 
and the image reconstructed by our approach with the CL (\ie\ w/ CL).
We observed that when using the CL,
our approach always achieves a better reconstruction quality; 
See the 3rd (and 7th) column versus the 4th (and 8th) column.
Besides, we discovered that our approach (w/o CL), 
in some cases, reconstructs adversarial examples 
with poor quality; See the 3rd and 7th columns in the last row.
By contrast, our method (w/ CL) still achieves a high reconstruction quality on these examples;
See the 4th and 8th columns in the last row.
These results demonstrate that our proposed loss can boost the performance
when learning visual representations from adversarial examples.

Next, we conduct experiments on CIFAR-10 
for quantitatively evaluating visual representations
by using the linear probing accuracy.
Specifically, we consider the standard accuracy as well as the robustness 
under the attack of $\text{A}^{3}$ and $\text{PF-A}^{3}$.
Table~\ref{tab:as-cl} presents the experimental results.
We observed that by utilizing the contrastive loss,
our approach achieves performance improvement of $4.9 \%$, $6.1 \%$, and $8.4\%$
on standard accuracy, robustness against A$^{3}$, and robustness against PF-A$^{3}$, respectively.
These empirical results demonstrate the necessity and importance of our proposed loss for learning high-quality visual representations.


\begin{table} [!t] 
    \small
    \centering
    \setlength\tabcolsep{2 pt}
   \caption{Ablation studies on the classifier, including (a) the contrastive loss (CL) in pre-training stage and (b) the adaptive ensemble (AE) in fine-tuning stage
   }
   \vspace{-0.5 em}
   \begin{minipage}{0.24\textwidth}
    \centering
    \begin{tabular}{@{}c|c|cc@{}}
        \toprule
        \multirow{2}{*}{Method} & \multirow{2}{*}{\begin{tabular}[c]{@{}c@{}}Standard \\ Accuracy\end{tabular}} & \multicolumn{2}{c}{~~Robustness}       \\ \cmidrule(l){3-4} 
                                &                                                                               & \multicolumn{1}{c}{A$^{3}$} & PF-A$^{3}$ \\ \midrule
        w/o CL                     & 74.6                                                                       & \multicolumn{1}{c}{35.7}   & 34.2    \\ 
        w/ CL            & 79.5                                                                                 & \multicolumn{1}{c}{41.8}   & 42.6    \\ \bottomrule
    \end{tabular}
    \subcaption{Pre-training}
    \label{tab:as-cl}
\end{minipage}%
\begin{minipage}{0.24\textwidth}
    \centering
    \begin{tabular}{@{}c|c|cc@{}}
        \toprule
        \multirow{2}{*}{Method} & \multirow{2}{*}{\begin{tabular}[c]{@{}c@{}}Standard \\ Accuracy\end{tabular}} & \multicolumn{2}{c}{~~Robustness}       \\ \cmidrule(l){3-4} 
                                &                                                                               & \multicolumn{1}{c}{A$^{3}$} & PF-A$^{3}$ \\ \midrule
        w/o AE                  & 81.9                                                                          & \multicolumn{1}{c}{38.9}   & 39.4    \\ 
        w/ AE                   & 90.3                                                                          & \multicolumn{1}{c}{44.5}   & 44.7    \\ \bottomrule
    \end{tabular}
    \subcaption{Fine-tuning}
    \label{tab:as-ae}
\end{minipage}
\vspace{-1.0 em}
\end{table}

\par\smallskip\noindent
{\bf Fine-tuning: Adaptive Ensemble.}
  Here, we conduct experiments to show the impact of our adaptive ensemble on the standard accuracy
  and the robustness.
  Table~\ref{tab:as-ae} lists the experimental results with/without our adaptive ensemble.
  Note that we employ the naive average ensemble when conducting experiments without our adaptive ensemble.
  From Table~\ref{tab:as-ae}, we observed that our adaptive ensemble significantly benefits the standard accuracy,
  with $8.4 \%$ performance improvement.
  Meanwhile, it boosts adversarial robustness against A$^{3}$ by $5.6 \%$ and against PF-A$^{3}$ by $5.3\%$.
  This is because the adaptive factor $p$ estimated by our detector
  can adaptively adjust the proportion of visual representations from clean and adversarial encoders,
  thereby significantly boosting the classification performance.

  \par\smallskip\noindent
  {\bf Fine-tuning: Masking Ratio.}
  We conduct experiments on CIFAR-10 to 
  explore how different masking ratios
  affect the performance of our approach during the finetuning.
  $12$ groups of masking ratios are taken into account, 
  ranging from $25 \%$ to $80 \%$.
  Note that in the pre-training,
  we directly set the masking ratio to $75 \%$ 
  by following MAE~\cite{he:cvpr22:mae};
  hence, no similar ablation study is required.
  Here, we consider the trade-off between standard accuracy and robustness
  (under A$^{3}$ and PF-A$^{3}$ attacks).

\begin{figure} [!t] 
    \centering
    \captionsetup[subfigure]{justification=centering}
    \begin{subfigure}[t]{0.23\textwidth}
        \centering
         \includegraphics[width=\textwidth]{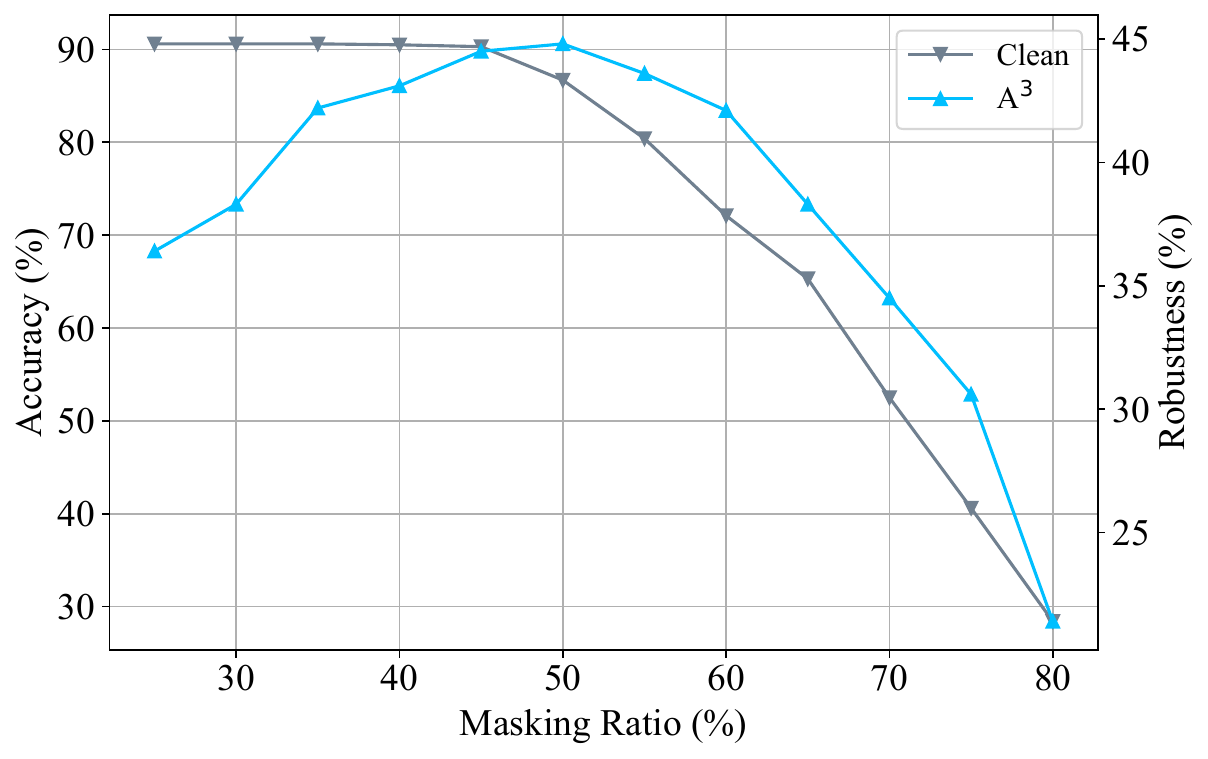}
          \vspace{-1.0 em}
        \caption{ Under A$^{3}$ attack
        }
        \label{fig:exp-mask-ratio-dense}
    \end{subfigure}
    \centering
     \begin{subfigure}[t]{0.23\textwidth}
        \centering
         \includegraphics[width=\textwidth]{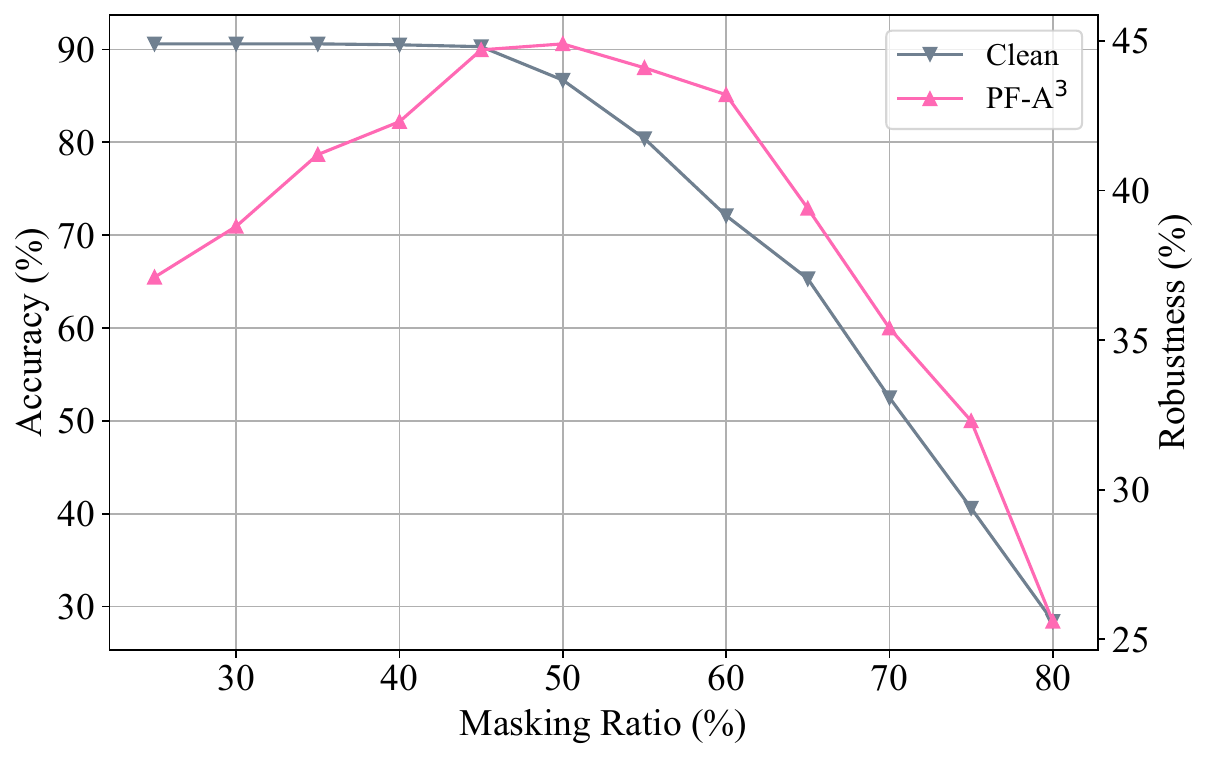}
           \vspace{-1.0 em}
        \caption{Under PF-A$^{3}$ attack
        }
        \label{fig:exp-mask-ratio-sparse}
    \end{subfigure}
  \vspace{-0.5em}
    \caption{
      Illustration of how different masking ratios in the finetuning affect the performance.
    }
    \label{fig:exp-mask-ratio}
    \vspace{-1.0 em}
\end{figure}

  Figures~\ref{fig:exp-mask-ratio-dense} and \ref{fig:exp-mask-ratio-sparse}
  illustrate experimental results.
  In Figure~\ref{fig:exp-mask-ratio-dense},
  we observed that increasing the masking ratio negatively affects
  standard accuracy (\ie\ the grey line) in all scenarios.
  In contrast, when the masking ratio is small (\ie\ $\leq 50\%$),
  a larger masking ratio benefits robustness against A$^{3}$ (\ie\ the blue line).
  But when the masking ratio is greater than $50 \%$,
  increasing the masking ratio hurts this robustness.
  This is because a small subset of masked patches  
  can eliminate the adversarial effect of adversarial attacks,
  while a large subset of masked patches would prevent our classifier from accurate classification.
  Clearly, our approach achieves the best trade-off on the masking ratio of $45 \%$,
  with standard accuracy of $90.3 \%$ and robustness against A$^{3}$ of $44.5 \%$.

  Similarly, Figure~\ref{fig:exp-mask-ratio-sparse} depicts robustness against
  PF-A$^{3}$~(\ie\ the pink line) under different masking ratios.
  We also include standard accuracy 
  (similar to Figure~\ref{fig:exp-mask-ratio-dense})
  for a better illustration of the trade-off.
  Obviously, when the masking ratio equals $45 \%$,
  our approach achieves the best trade-off,
  with standard accuracy of $90.3\%$
  and robustness of $44.7 \%$.
  Based on the above discussion, we set our masking ratio to $45 \%$
  to ensure the best trade-off between standard accuracy
  and robustness (under adaptive attacks).

\subsection{Evaluating Our Detector}
\label{sec:exp-det}

In this section, we conduct experiments on CIFAR-10
for comparing our detector with four detection baselines,
\ie\ \textbf{Odds}~\cite{roth:icml19:odds}, \textbf{NIC}~\cite{ma:ndss19:nic}, 
\textbf{GAT}~\cite{yin:iclr20:gat},
and~\textbf{JTLA}~\cite{raghuram:icml21:jtla}.
Three aforementioned  adaptive attacks
under two small attack constraints,
\ie\ $\epsilon=2/255$ and $\epsilon=4/255$, 
are used for evaluating detection accuracy. 
Table~\ref{tab:det} lists the detection accuracy values under different attack methods.
We observed that our detector achieves the best detection accuracy under all scenarios.
Specifically, our approach achieves the best detection accuracy of $99.4 \%$ 
under the attack constraint of $\epsilon = 4/255$ (see the 5th column).
Decreasing the attack constraint to $2/255$ increases the detection difficulty,
with our approach still maintaining the superior detection accuracy of $95.8 \%$ (see the 4th column) in the worst case.
Besides, our detector outperforms all baselines,
with the detection accuracy improvements ranging from $1.8 \%$ (\ie\ $95.9 \%$ vs. $94.1 \%$, see the 3rd column) 
to $6.3 \%$ (\ie\ $96.4 \%$ vs. $90.1 \%$, see the 2nd column).
The statistical evidence exhibits that our two new designs for the detector,
\ie\ the new Multi-head Self-Attention (MSA) mechanism 
and the proposed loss function, 
are effective for exposing adversarial perturbation, 
rendering our detector to better defend against adaptive attacks.


\begin{table} [!t] 
    \small
    \centering
    \setlength\tabcolsep{2 pt}
   \caption{Comparisons of detection accuracy on CIFAR-10 under different adaptive attacks
   with best results shown in bold
   }
   \vspace{-0.5 em}
   \begin{tabular}{@{}c|ccc|ccc@{}}
    \toprule
    \multirow{2}{*}{Method} & \multicolumn{3}{c|}{Attack Constraint ( $ \epsilon = 2 / 255$ )}             & \multicolumn{3}{c}{Attack Constraint ( $ \epsilon = 4 / 255 $)}             \\ \cmidrule(l){2-7} 
                            & \multicolumn{1}{c}{AutoAttack} & \multicolumn{1}{c}{A$^{3}$}   & PF-A$^{3}$ & \multicolumn{1}{c}{AutoAttack} & \multicolumn{1}{c}{A$^{3}$}   & PF-A$^{3}$                         \\ \midrule
    Odds                    & \multicolumn{1}{c}{90.1}       & \multicolumn{1}{c}{90.6} & 91.2  & \multicolumn{1}{c}{94.8}       & \multicolumn{1}{c}{94.3} & 94.9  \\ 
    NIC                     & \multicolumn{1}{c}{93.1}       & \multicolumn{1}{c}{92.5} & 94.4  & \multicolumn{1}{c}{95.8}       & \multicolumn{1}{c}{95.6} & 96.4  \\ 
    GAT                     & \multicolumn{1}{c}{92.6}       & \multicolumn{1}{c}{93.8} & 93.0    & \multicolumn{1}{c}{96.0}         & \multicolumn{1}{c}{95.8} & 95.6  \\ 
    JTLA                    & \multicolumn{1}{c}{94.3}       & \multicolumn{1}{c}{94.1} & 93.9  & \multicolumn{1}{c}{95.6}       & \multicolumn{1}{c}{96.4} & 96.2  \\ \midrule
    Ours &
      \multicolumn{1}{c}{\textbf{96.4}} &
      \multicolumn{1}{c}{\textbf{95.9}} &
      \textbf{95.8} &
      \multicolumn{1}{c}{\textbf{99.4}} &
      \multicolumn{1}{c}{\textbf{98.7}} &
      \textbf{98.9} \\ \bottomrule
    \end{tabular}
   \label{tab:det}
   \vspace{-1.0 em}
\end{table}


\begin{table} [!t] 
    \small
    \centering
    \setlength\tabcolsep{15 pt}
   \caption{Ablation studies on our detector
   }
   \vspace{-1.0 em}
    \begin{tabular}{@{}c|cc@{}}
        \toprule
        Method      & A$^{3}$ & PF-A$^{3}$ \\ \midrule
        w/o GB      & 92.3   & 91.9    \\ 
        w/o MSA      & 92.9   & 92.8    \\ \midrule
        Ours       & 98.7   & 98.9    \\ \bottomrule
        \end{tabular}
   \label{tab:exp-ab-det}
   \vspace{-1.5 em}
\end{table}

\subsection{Ablation Studies on Our Detector}
\label{sec:as-det}


%
\par\smallskip\noindent
{\bf MSA on the Detection Accuracy.}
Here, we empirically show how our developed MSA mechanism affects 
the detection accuracy under the attack of A$^{3}$ and PF-A$^{3}$.
We consider two scenarios.
First, we remove the Guided Backpropagation (GB) variant 
to validate whether it benefits the detection of adversarial examples, 
denoted as ``w/o GB''.
Second, we discard our proposed MSA and instead naively add two sets of patch embeddings
respectively from the clean image and the GB variant, 
denoted as ``w/o MSA''.
Table~\ref{tab:exp-ab-det} lists the experimental results.
We discovered that simply adding two sets of patch embeddings only marginally improves the detection accuracy
of $0.6 \%$ (or $0.9 \%$) under the A$^{3}$ (or PF-A$^{3}$) attack (see ``w/o GB'' vs. ``w/o MSA'').
Equipped with our MSA mechanism, in sharp contrast,
the Guided Backpropagation technique can significantly benefit the detection task,
with the detection accuracy improvement
of $6.4 \%$ (or $7.0 \%$) under the A$^{3}$ (or PF-A$^{3}$) attack (see ``w/o GB'' vs. ``Ours'').
These results confirm that
(\romsm{1}) the Guided Backpropagation technique can help expose adversarial perturbation
and (\romsm{2}) our proposed MSA can significantly boost the detector's robustness against adaptive attacks.

\begin{figure}[!t]
  \captionsetup[subfigure]{justification=centering}
  \begin{subfigure}[t]{0.23\textwidth}
      \centering
       \includegraphics[width=\textwidth]{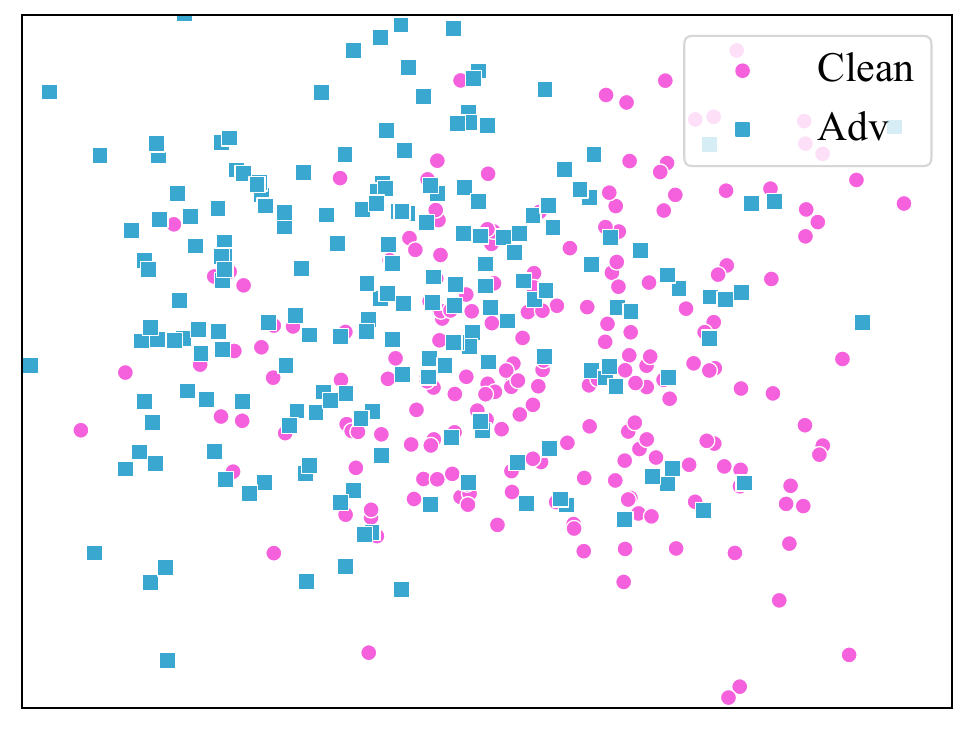}
         \vspace{-1.5 em}
      \caption{w/o SNN loss \\ (under A$^{3}$ attack)
      }
      \label{fig:exp-vit-as-det-a3}
  \end{subfigure}
  \centering
   \begin{subfigure}[t]{0.23\textwidth}
      \centering
       \includegraphics[width=\textwidth]{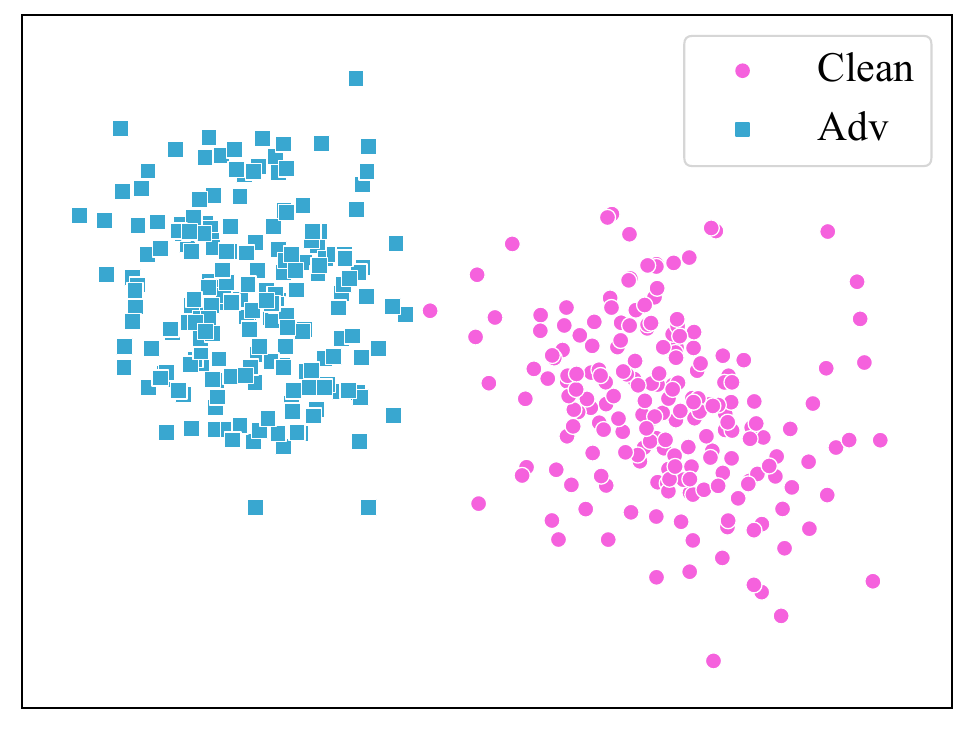}
         \vspace{-1.5 em}
      \caption{w/ SNN loss \\ (under A$^{3}$ attack)
      }
      \label{fig:exp-vit-as-det-ours-a3}
  \end{subfigure}
  \centering
  \begin{subfigure}[t]{0.23\textwidth}
     \centering
      \includegraphics[width=\textwidth]{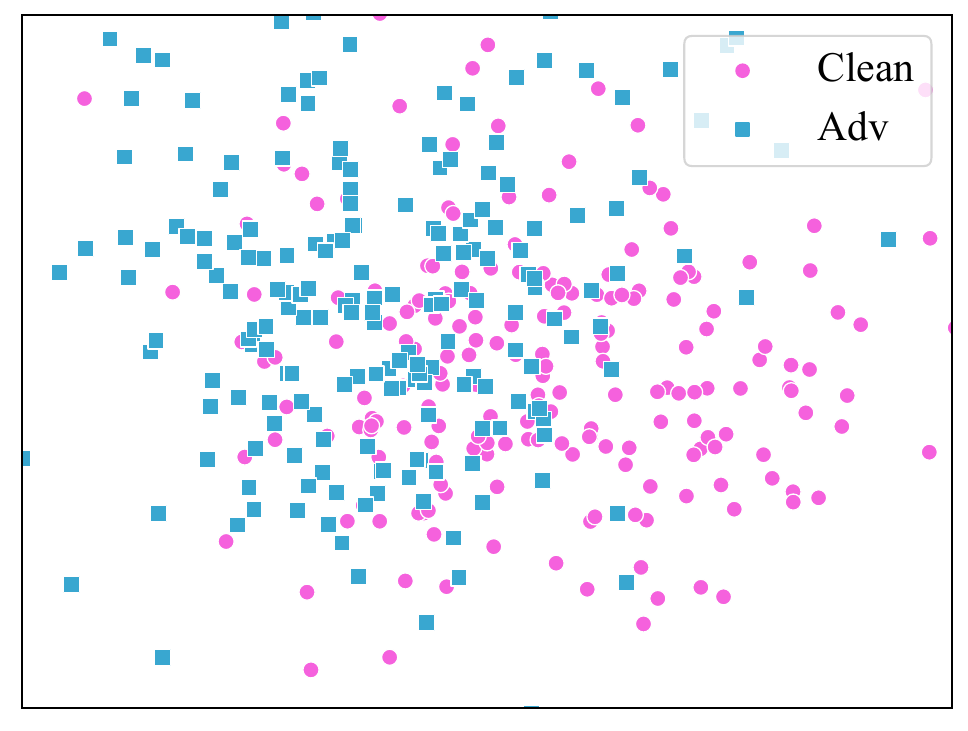}
        \vspace{-1.5 em}
     \caption{ w/o SNN loss \\ (under PF-A$^{3}$ attack)
     }
     \label{fig:exp-vit-as-det-pf-a3}
  \end{subfigure}
  \centering
  \captionsetup[subfigure]{justification=centering}
  \begin{subfigure}[t]{0.23\textwidth}
      \centering
       \includegraphics[width=\textwidth]{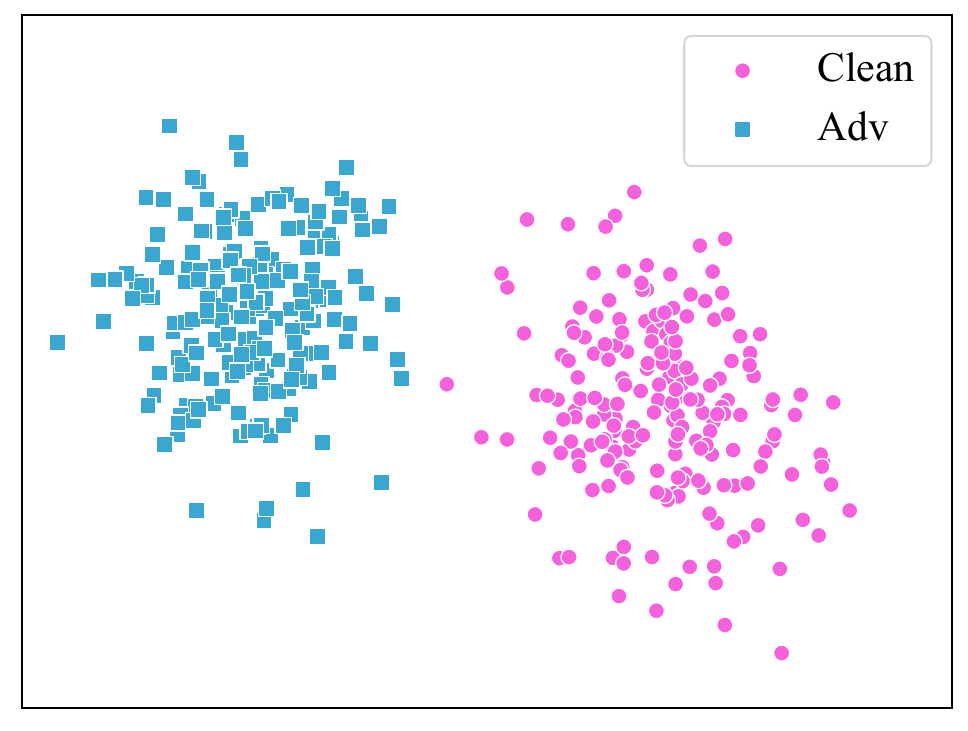}
         \vspace{-1.5 em}
      \caption{w/ SNN loss \\ (under PF-A$^{3}$ attack)
      }
      \label{fig:exp-vit-as-det-ours-pf-a3}
  \end{subfigure}
\vspace{-0.5 em}
  \caption{
    t-SNE visualization on CIFAR-10 by using our detector with/without SNN loss.
    For each experiment, we perform t-SNE visualization on $200$ clean images
    and $200$ adversarial examples 
    generated either by the A$^{3}$ attack, \ie\ (a) and (b), 
    or by the PF-A$^{3}$ attack, \ie\ (c) and (d).
  }
  \label{fig:exp-as-det}
  \vspace{-1.5 em}
\end{figure}

\par\smallskip\noindent
{\bf SNN Loss on Visual Representations.}
Here, we reveal the effect of our proposed loss, \ie\ \eqref{eq:det-loss}, on detecting adversarial examples.
We consider how our detector with or without the Soft-Nearest Neighbors (SNN) loss 
affects the resulting representation space.
In particular, we employ t-SNE visualization~\cite{maaten:jmlr08:t-sne}
on $200$ clean images randomly sampled from CIFAR-10
and $200$ adversarial examples generated either by the A$^{3}$ attack
or by the PF-A$^{3}$ attack.
Figures~\ref{fig:exp-vit-as-det-a3} and \ref{fig:exp-vit-as-det-ours-a3} depict
the results by using the A$^{3}$ attack, 
while Figures~\ref{fig:exp-vit-as-det-pf-a3} and \ref{fig:exp-vit-as-det-ours-pf-a3}
present the results by employing the PF-A$^{3}$ attack.
We observed that without the SNN loss, 
the representations for clean images and adversarial examples are highly entangled;
see Figures~\ref{fig:exp-vit-as-det-a3} and ~\ref{fig:exp-vit-as-det-pf-a3}.
In sharp contrast, by minimizing the SNN loss,
the representations for clean images and adversarial examples are mutually isolated,
as shown in Figures~\ref{fig:exp-vit-as-det-ours-a3} and \ref{fig:exp-vit-as-det-ours-pf-a3},
making adversarial examples detectable.

\section{Conclusion}
\label{sec:conslusion}

This article has proposed a novel Vison Transformers (ViT) architecture, including a detector and a classifier,
which are bridged by a newly developed adaptive ensemble.
This ViT architecture enables us to boost adversarial training to defend against adaptive attacks, 
and to achieve a better trade-off between standard accuracy and robustness.
Our key idea includes introducing a novel Multi-head Self-Attention (MSA) mechanism 
to expose adversarial perturbations for better detection
and employing two decoders to extract visual representations respectively from clean images and adversarial examples so as to reduce the negative effect of adversarial training on standard accuracy.
Meanwhile, our adaptive ensemble lowers potential adversarial effects upon encountering adversarial examples
by masking out a random subset of image patches across input data.
Extensive experiments have been conducted for evaluation, 
showing that our solutions significantly outperform their state-of-the-art counterparts
in terms of standard accuracy and robustness.

\section*{Acknowledgments}
This work was supported in part by NSF under Grants 2019511, 2348452, and 2315613. Any opinions and findings expressed in the paper are those of the authors and do not necessarily reflect the views of funding agencies.

\bibliographystyle{ACM-Reference-Format}
\balance
\bibliography{main}


\end{document}